\documentclass[10pt,journal,compsoc]{IEEEtran}

\usepackage{times}
\usepackage{epsfig}
\usepackage{rotating,graphicx}
\usepackage{amsmath}
\usepackage{amssymb}
\usepackage{dblfloatfix}
\usepackage{algorithm}
\usepackage{fixltx2e}
\usepackage{mathtools}
\usepackage[noend]{algpseudocode}
\usepackage{supertabular,booktabs}
\usepackage{subcaption}
\usepackage{placeins}
\usepackage{flafter}
\usepackage{enumitem}
\usepackage{multirow}
\usepackage{booktabs}
\usepackage{color}
\usepackage{url}
\usepackage{soul}

\usepackage[dvipsnames]{xcolor}

\newcommand{\etal}{\textit{et al}.}

\ifCLASSOPTIONcompsoc
  \usepackage[nocompress]{cite}
\else

  \usepackage{cite}
\fi

\ifCLASSINFOpdf

\else

\fi

\begin{document}

\title{A Comparative Evaluation of SGM Variants (including a New Variant, tMGM) for Dense Stereo Matching}

\author{\IEEEauthorblockN{Sonali Patil \IEEEauthorrefmark{1},
Tanmay Prakash \IEEEauthorrefmark{2},
Bharath Comandur \IEEEauthorrefmark{3}, and
Avinash Kak \IEEEauthorrefmark{4}
}\\
\IEEEauthorblockA{School of Electrical and Computer Engineering, Purdue University\\
West Lafayette, IN, USA  \\ \IEEEauthorrefmark{1} patil19@purdue.edu \quad \IEEEauthorrefmark{2}  tprakash@purdue.edu \quad \IEEEauthorrefmark{3} bcomandu@purdue.edu \quad \IEEEauthorrefmark{4} kak@purdue.edu} }

\author{Sonali Patil,
        Tanmay Prakash,
        Bharath Comandur, and 
        Avinash Kak
\IEEEcompsocitemizethanks{\IEEEcompsocthanksitem  The authors are with the Robot Vision Laboratory, School of Electrical and Computer Engineering, Purdue University, West Lafayette, IN 47907. \protect\\
E-mail: \{patil19,tprakash, bcomandu,kak\}@purdue.edu
}
\thanks{}}

\markboth{}%
{Shell \MakeLowercase{\textit{et al.}}: A Comparative Evaluation of SGM Variants (including a New Variant, tMGM) for Dense Stereo Matching}

\maketitle

\newcommand{\unclear}[1]{\textcolor{magenta}{\textsf{\emph{\textcolor{magenta}{#1}}}}}
\newcommand{\new}[1]{\textcolor{blue}{\textsf{\emph{\textcolor{blue}{#1}}}}}
\definecolor{darkblue}{RGB}{0, 0, 200}

\begin{abstract}
  Our goal here is threefold: [1] To present a new dense-stereo
  matching algorithm, tMGM, that by combining the hierarchical logic
  of tSGM with the support structure of MGM achieves 6-8\% performance
  improvement over the baseline SGM (these performance numbers are
  posted under tMGM-16 in the Middlebury Benchmark V3 ); and [2]
  Through an exhaustive quantitative and qualitative comparative
  study, to compare how the major variants of the SGM approach to
  dense stereo matching, including the new tMGM, perform in the
  presence of: (a) illumination variations and shadows, (b) untextured
  or weakly textured regions, (c) repetitive patterns in the scene in
  the presence of large stereo rectification errors. [3] To present a
  novel DEM-Sculpting approach for estimating initial disparity search
  bounds for multi-date satellite stereo pairs. Based on our study, we
  have found that tMGM generally performs best with respect to all
  these data conditions.  Both tSGM and MGM improve the density of
  stereo disparity maps and combining the two in tMGM makes it
  possible to accurately estimate the disparities at a significant
  number of pixels that would otherwise be declared invalid by
  SGM. The datasets we have used in our comparative evaluation include
  the Middlebury2014, KITTI2015, and ETH3D datasets and the satellite
  images over the San Fernando area from the MVS Challenge dataset.
\end{abstract}

\IEEEpeerreviewmaketitle

\section{Introduction}

Many modern matching methods for calculating dense disparity
maps from stereo pairs are rooted in Markov Random Field
(MRF) modeling of the disparity maps, which allows for the
joint probability distribution of the disparity values over
a reference image to be expressed as a product of the
potentials over local neighborhoods. This simplified
representation of the disparity probability distributions
allows an ``energy'' function to be defined whose
minimization leads to a MAP estimate of the disparities. The
energy function consists of two ``costs'': the first
represents the cost associated with the assignment of
disparities to the individual pixel positions in the
reference image and the second represents the cost
associated with the assignment of two different disparities
at two neighboring pixel positions.  The first cost is
frequently referred to as the Data Cost and the second as
the Discontinuity Cost.  An immediate consequence of MRF
modeling is that the second cost only involves local
neighborhoods around each pixel position in the reference
image.
\begin{figure}
\begin{center}
\includegraphics[scale=0.6]{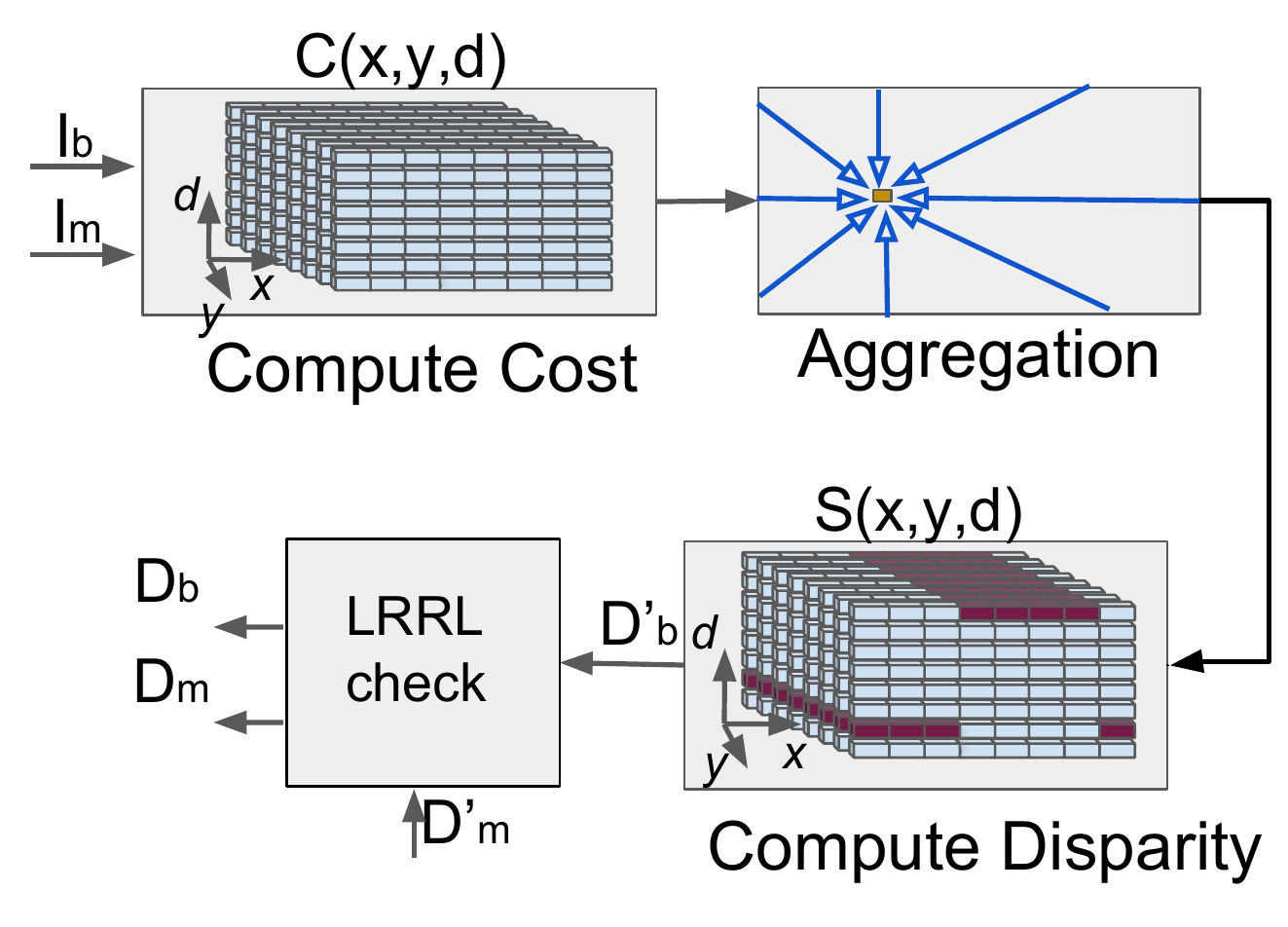}
\end{center}
\caption{The SGM algorithm is applied to a rectified image
  pair $(\textbf{I}_b, \textbf{I}_m)$ as input where
  $\textbf{I}_b$ is the base image and $\textbf{I}_m$ the
  match image.  (In the narrative, we have also referred to
  the base image as the reference image.)  The SGM pipeline
  starts out by computing the Data Cost by using the Census
  Transform at each pixel for each disparity $d$ in a given
  disparity range $[d_{min},d_{max}]$.}
\label{fig:SGM_overview}
\end{figure}

While it is relatively straightforward to construct a
theoretical formalism along the lines indicated above and to
end up with an energy function whose minimization would lead
to a solution for the disparities, it is an entirely
different matter to come up with a viable computational
approach for the minimization of the energy function ---
especially in light of the fact that solving the energy
minimization problem exactly can be shown to be NP-Hard
\cite{boykov2001fast}.  What that means is that we can only
construct approximate solutions to the minimization of the
energy function. The question then becomes as to whether the
quality of the resulting solutions is acceptable.

Fortunately, with the advent of SGM (Semi-Global Matching)
based solutions, as first advanced by Hirschm\"{u}ller
\cite{hirschmuller2008stereo}, we now have approximate
solutions that are not only computationally efficient but
that also produce high quality disparity maps. SGM
aggregates the Data and the Discontinuity costs along
several symmetric 1D paths for each possible disparity at
each pixel position in the reference image. Subsequently,
the summed energy at each pixel position is calculated by
simply adding up the contributions calculated along each of
the paths.  Finally, in a winner-take-all strategy, we
retain at each pixel that disparity which has associated
with it the minimum summed energy.

Over the years, several authors have proposed modifications
to SGM to improve the quality of the results obtained. 
Most of the variants include one or more of the following:
\begin{enumerate}[topsep=0pt,itemsep=-1ex,partopsep=1ex,parsep=1ex]
\item Using different data cost terms \cite{Zbontar2016}
\item Adapting the penalty coefficients in the
  discontinuity-preserving smoothness term to the gradients
  (\cite{rothermel2012sure},\cite{hirschmuller2008stereo}),
  or learning them directly from the data \cite{Seki2017}
\item Using different aggregation strategies \cite{facciolo2015mgm}
\item Truncating the disparity search range using
  hierarchical approaches \cite{rothermel2012sure}
\end{enumerate}
Of the several variants that now exist, MGM by Facciolo et
al. \cite{facciolo2015mgm} and tSGM by Rothermel
\cite{rothermel2012sure} are arguably among the most
prominent. MGM was motivated by the perceived shortcomings
associated with scanline based support regions in SGM
whereas tSGM was developed to reduce the memory and runtime
demands of SGM. In MGM, the scanline based star-shaped
support regions are replaced by ``quad-area'' based support
regions that allow neighboring pixel positions to exert
greater influence on one another's disparity values.  In
tSGM, the processing is performed in a coarse-to-fine manner
that allows a coarse level to constrain the disparity range
that must be searched at the next finer level --- which
results in substantial savings in computation.
Additionally, in tSGM, using the same smoothness parameters
and the same matching criterion for the corresponding pixels
(specifically the Census Filter) at all levels in the
hierarchy enables the coarser levels to act as ``mediators''
for suppressing the noise and the artifacts in the finer
levels.

It should therefore be no surprise that we would want to
pull together the benefits of MGM and tSGM, which is exactly
what we have done in a new dense stereo matching algorithm
that, out of respect for its origins, we have named tMGM.

That has dictated the following two goals for the present
submission: First, to propose the tMGM alternative for dense
stereo matching.  And, second, to provide comparative
insights into how the different SGM variants, including the
new tMGM, differ with regard to the matching difficulties
created by the different types of scene conditions:
illumination differences between the two images of a stereo
pair, the presence of untextured or weakly textured regions,
and the presence of repetitive patterns especially when
there exist large stereo rectification errors.

With regard to how the new approach of tMGM fares vis-a-vis
the more established approaches, our results show that it is
more robust to challenging scene conditions compared to its
competitors and is suitable for large scale applications
involving satellite images.


Our comparative evaluation is based on the following four
datasets: Middlebury2014, KITTI2015, ETH3D, and the
satellite images over the San Fernando area that were used
for the MVS challenge \cite{MVS_Challenge}.  This evaluation
provides both quantitative and qualitative support for the
main conclusions drawn in this paper.

For the Middlebury2014, KITTI2015, and ETH3D datasets, the
quantitative support is based on the {\em invalid pixel
  error}, {\em the bad pixel error}, {\em the total error},
and {\em the average error} metrics.  For the satellite
images, the ground truth is provided in the form of 2.5D
LiDAR data that can be compared with the Digital Surface
Models (DSMs) constructed from the stereo pairs. This
comparison yields a quantitative evaluation using the {\em
  completeness}, {\em {median error}} and {\em {RMSE}}
metrics as defined in \cite{MVS_Challenge}.

Our overall conclusion is that the hierarchical variants of
the SGM algorithm perform better than their non-hierarchical
counterparts for all cases represented by the stereo pairs
in all four datasets we have used in this study.  As we will
show, the sensitivity of the SGM variants to important
confounding factors varies both as a function of the variant
itself and as a function of the support structure. By
confounding-factors we obviously mean the illumination
differences between the images, the presence of untextured
or weakly textured regions in the images, and the presence
of repetitive patterns (especially when there exist large
rectification errors in the regions with repetitive
patterns).

We now list here some of the specifics in the conclusions we
have drawn from our study. In the listing shown below, we
have used the notation xGM-$N$, where x stands for ``S'',
``M'', ``tS'', or ``tM'' and the integer $N$ for the value
of the main parameter that controls the shape of the support
region.

\begin{itemize}
\item In the presence of significant illumination
  differences in a stereo pair, the tMGM-8 and the tMGM-16
  algorithms are the best to use for disparity
  calculations. Illumination differences may be caused by
  the shadows cast by the objects in a scene, by the
  presence of pixels that correspond to specularly
  reflecting scene points, and, even more ordinarily, by the
  different lighting conditions under which the two images
  of a stereo pair were recorded (something that is likely
  to be the case on account of the different sun angles for
  out-of-date satellite images).

\item If untextured or weakly textured regions dominate the
  scene, we are likely to get the most accurate disparity
  maps with the proposed tMGM-16 algorithm.

\item If a scene is rich in repetitive structures and also
  has large rectification errors in such regions, we are
  likely to get the best disparity results with either the
  tSGM-8 or the tMGM-8 algorithm.
\end{itemize}

The rest of the paper is organized as follows: In Section 2
we quickly review the related work. Section 3 then briefly
describes each of the main algorithms --- SGM, MGM, tSGM,
and tMGM. In Section 4, we present our comparative results
for the four different algorithms and for different
parameter choices related to the shape of the support
region. In Section 5, we discuss further refinements to the tMGM
algorithm and evaluation on the Middlebury Benchmark V3. We
finally conclude in Section 6.

\section{Related Work}

The dense stereo correspondence problem is fundamental to
many larger problems in robotics, remote sensing,
virtual/augmented reality, etc., and as such has
inspired a diverse set of algorithms. Scharstein and
Sziliski \cite{Scharstein2002} created a taxonomy by which
to understand them all, dividing the approaches into global,
local, cooperative, and dynamic programming approaches.

Global methods, which model the entire disparity map with a
2D Markov Random Field as described in the introduction, are
the primary interest of this paper. As mentioned earlier,
though calculating the MAP estimate of a 2D MRF is NP-Hard,
approximate algorithms have produced promising results. The
work presented in \cite{boykov2001fast} uses graph-cut to
find a local minimum within a known factor of the global
minimum. Drory \etal \cite{drory2014semi} show that SGM can
be understood as a belief propagation based approach to
optimizing the joint probability defined by the 2D MRF. In
this interpretation however, it is shown that
the cost term is overcounted and \cite{drory2014semi}
proposes an overcounting correction.

There is a significant body of recent work that attempts to
use machine learning, often deep-learning, to solve the
dense stereo correspondence problem. Many contributions,
however, still utilize semi-global matching in some form,
for instance as a refinement of initial disparities
\cite{Zbontar2016} or as the model for which the parameters
are learned \cite{Seki2017}. As such, a study of the
underlying parameters of SGM and its variants may further
improve these approaches.

The other approaches in the taxonomy presented by Scharstein
and Sziliski \cite{Scharstein2002} generally avoid directly
addressing the global distribution of disparity. Local
methods consider the neighborhood of intensity values,
rather than disparity values, simplifying computation
\cite{Kanade1994,Yoon2006}. Dynamic programming approaches
consider individual scanlines, since 1D MRFs can be
optimized in polynomial time \cite{Birchfield1999}, and
often rely on post-processing to reduce the inconsistency
between adjacent scanlines. Cooperative algorithms rely on
iterating over local operations until convergence produces
global order \cite{Marr1976}. 

Public challenges like the Middlebury Stereo Challenge
\cite{Scharstein2002} and the KITTI Vision Benchmark
\cite{Menze2018JPRS} provide comparisons of a number of
different stereo algorithms. While these challenges serve to
determine which algorithms perform the best on a set of
metrics, this paper attempts to delve deeper into how
varying the underlying parameters affects performance. Note
that the paper by D'Angelo \cite{DAngelo2016} also presents
a comparison of SGM-8, SGM-16, ocSGM-16 (oc - overcounting
corrected), and MGM-8 on satellite image data, but does not
include MGM-16, tSGM, or tMGM.

\section{SGM and Its Variants}

The MRF based approach to the estimation of a univalued
disparity over the pixel positions in the reference image of
a rectified stereo pair is stated through the minimization
of the following expression for ``energy'':
\begin{align}
\mathcal{E}(\textbf{D}) &= \sum_{\textbf{p}} \Big( \mathcal{C}(\textbf{p}, \textbf{D}_\textbf{p}) + \sum_{\textbf{q}\in N_\textbf{p} } \mathcal{V} (\textbf{D}_\textbf{p}, \textbf{D}_\textbf{q}) \Big)
\label{eq0}
\end{align}
where $\textbf{p} = [\textbf{p}_x,\textbf{p}_y]^T$
represents a pixel position in the reference image of a
stereo pair and $\textbf{D}_\textbf{p}$ the disparity
assigned to that pixel position. 

The notation $\mathcal{C}(\textbf{p},
\textbf{D}_\textbf{p})$, referred to as the Data Cost,
involves comparing the reference image pixel at $\textbf{p}$
with the other image pixel at
$[\textbf{p}_x+\textbf{D}_\textbf{p},\textbf{p}_y]^T$. The
notation $\mathcal{V}
(\textbf{D}_\textbf{p},\textbf{D}_\textbf{q})$, called the Discontinuity Cost,
denotes

the cost of assigning two different disparities to two pixel
positions $\textbf{p}$ and $\textbf{q}$ where the latter is
in the neighborhood $N_{\textbf{p}}$ associated with the former.

We want to find the disparity map that minimizes the energy
function shown above.

For stereo matching, the expression for energy is
expressed in the following manner \cite{hirschmuller2008stereo}:
\begin{align}
\mathcal{E}(\textbf{D}) &= \sum_{\textbf{p}} \Big( \mathcal{C}(\textbf{p}, \textbf{D}_\textbf{p}) + \sum_{\textbf{q}\in N_\textbf{p} } P_1 \mathcal{T}[|\textbf{D}_\textbf{p}  -\textbf{D}_\textbf{q}|=1] \notag \\  & + \sum_{\textbf{q}\in N_\textbf{p} } P_2 \mathcal{T}[|\textbf{D}_\textbf{p}-\textbf{D}_\textbf{q}|>1] \Big) 
\label{eq1}
\end{align}
Now the Discontinuity Cost, at pixel $\textbf{p}$, is broken
into two separate parts, one for the case when the disparity
values at $\textbf{p}$ and $\textbf{q}$ differ by exactly 1
and, two, when they differ by more than 1.  The notation
$\mathcal{T}[\cdot]$ evaluates the truthvalue of the
predicate.  The two portions of the Discontinuity Cost carry
the user-supplied weights $P_1$ and $P_2$.  Since the
minimization of the Discontinuity Cost is to ensure local
smoothness of the reconstructed surfaces, we assign a small
penalty $P_1$ to the case when the disparity varies by 1 for
pixels in the neighborhood $N_\textbf{p}$ of pixel
$\textbf{p}$. This term takes care of slanted or gently
curved surfaces.  And we assign a larger penalty $P_2$ when
the disparity changes more rapidly.
 
 \begin{figure*}
\begin{center}
\includegraphics[width=\linewidth]{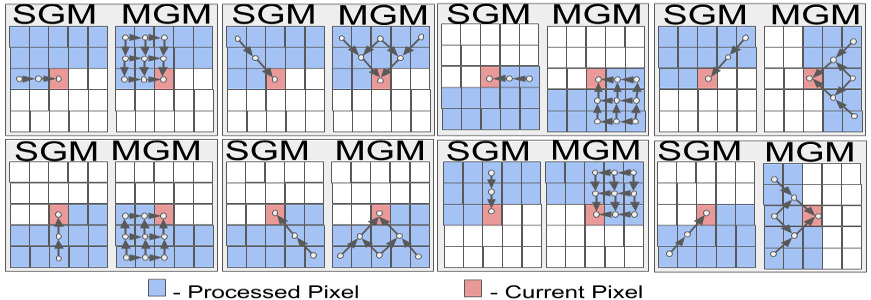}
\end{center}
   \caption{Scanning directions and support coverage for
     SGM-8 and MGM-8.  Whereas SGM provides for line-based
     coverage through a set of scanlines arranged in a star
     pattern, MGM provides for ``quad-area'' based coverage
     in the manner shown.}
\label{fig:MGM}
\end{figure*}
\subsection{Baseline SGM}
Unfortunately, the 2D global energy minimization as given in
Eq. (\ref{eq1}) is NP-Hard \cite{boykov2001fast}.  In the
baseline SGM implementation \cite{hirschmuller2008stereo},
we aggregate the cost recursively along several 1D
directions. The aggregated cost along a single direction
$\textbf{r}$ is given as follows:
\begin{align}
\mathcal{L}_{\textbf{r}} (\textbf{p},d) &= \mathcal{C}(\textbf{p},d)+\min \Big( \mathcal{L}_{\textbf{r}}(\textbf{p}-\textbf{r},d), \mathcal{L}_{\textbf{r}}(\textbf{p}-\textbf{r},d-1)+P_1, \notag \\ 
& \mathcal{L}_{\textbf{r}}(\textbf{p}-\textbf{r},d+1)+P_1, \min_{i} \mathcal{L}_{\textbf{r}}(\textbf{p}-\textbf{r},i)+P_2 \Big) \notag \\
\label{eq2}
\end{align}
where the notation
$\mathcal{L}_{\textbf{r}}(\textbf{p}-\textbf{r},d)$ means
that we are now calculating a line-based estimation of the
energy with the direction of the line given by the vector
$\textbf{r}$.  Note that $\mathcal{L}_{\textbf{r}}
(\textbf{p},d)$ is initialized with the data cost
$\mathcal{C}(\textbf{p},d)$ for all pixel positions
$\textbf{p}$ and all possible disparities $d$. Also, the
range of values spanned by the variable $i$ in the inner
minimization in Eq. (\ref{eq2}) is over all possible
disparities.  Comparing with Eq. (\ref{eq1}), the terms
$\mathcal{L}_{\textbf{r}}(\textbf{p}-\textbf{r},d \pm
1)+P_1$ are for the case of locally smooth disparity (i.e.,
when the disparity difference is $\pm 1$ along the direction
$\textbf{r}$). In this case, we use the penalty $P_1$ which
is relatively small.  On the other hand, when the disparity
differences are locally large, the last term, $\min_{i}
\mathcal{L}_{\textbf{r}}(\textbf{p}-\textbf{r},i)+P_2$,
kicks in. In this case, the penalty coefficient $P_2$ is
relatively large. To allow disparity discontinuities along
the edges in an image, $P_2$ can be adapted to the response
of an edge operator.  When a disparity map is not locally
smooth, that is, when the disparity differences exceed $\pm
1$, the outer minimization in Eq. (\ref{eq2}) will be
dominated by what is returned by the inner minimization that
depends on the penalty $P_2$.

\begin{algorithm}
\caption{Core SGM Algorithm} 
\label{algo:SGM}
\hspace*{\algorithmicindent} \textbf{Input:} $\textbf{I}_b$,$\textbf{I}_m$, $d_{min}$, $d_{max}$, $\textbf{r}$ \\
\hspace*{\algorithmicindent} \textbf{Output:} $\textbf{D}_b$,$\textbf{D}_m$
{\small   \begin{algorithmic}[1]
 \Procedure{Core SGM}{}
\State $\mathcal{C} = census\_cost(\textbf{I}_b,\textbf{I}_m, d_{min}, d_{max})$
\State $canny = canny\_filter(\textbf{I}_b)$
\State $\mathcal{S}= aggregate\_cost(\mathcal{C}, \textbf{r},canny )$
\State $\textbf{D}'_b = Compute\_disp(\mathcal{S})$
\State $Median\_filter(\textbf{D}'_b,3)$
\State $\mathcal{C} = census\_cost(\textbf{I}_m,\textbf{I}_b, -d_{max}, -d_{min})$
\State $canny = canny\_filter(\textbf{I}_m)$
\State $\mathcal{S}= aggregate\_cost(\mathcal{C},\textbf{r}, canny )$
\State $\textbf{D}'_m = Compute\_disp(\mathcal{S})$
\State $Median\_filter(\textbf{D}'_m,3)$
  \EndProcedure
 \end{algorithmic} }
    \end{algorithm}
The total aggregated cost along all the directions is given
as
\begin{align}
  \mathcal{S}(\textbf{p},d) &= \sum_{\textbf{r}} \mathcal{L}_{\textbf{r}}(\textbf{p},d)
  \label{eq:sum_eq}
\end{align}
A disparity map is computed from the $\mathcal{S}$ values by
taking the argmin of $\mathcal{S}$ at each pixel over all
possible disparities. Fig. \ref{fig:SGM_overview} and
Algorithm \ref{algo:SGM} summarize the steps for the core
SGM algorithm. 

The data term $\mathcal{C}$ is pre-calculated for the given
disparity range. Since the data term is prone to noise,
which may be due to the presence of untextured or weakly
textured regions, different radiometric characteristics of
the two images, and so on, some smoothing of the data term
is required.  Therefore, the aggregation step includes some
smoothing.  Aggregation is carried out along each horizontal
slice of $\mathcal{C}$, with each such slice corresponding
to one value of disparity, going from bottom to top in the
disparity volume. Note that, on account of the vertical
dependencies, the horizontal slices cannot be processed
independently and the bottom-to-top order within the
disparity volume is important. Aggregation along each
direction $\mathcal{L}_\textbf{r}$ results in a volume and
summing all these volumes gives us the final sum volume
$\mathcal{S}$. Disparity map $\textbf{D}'_b$ is calculated
by taking argmin over the sum volume $\mathcal{S}$. Since we
are aggregating along several directions and not only along
epipolar lines, we cannot detect occluded areas in this
framework. Therefore, a consistency check in the form of a
Left-Right-Right-Left (LRRL) check is carried out on the
disparity map over the base image $\textbf{D}'_b$ and on the
match image $\textbf{D}'_m$ in order to detect the invalid
pixels in the base image. Note that a pixel may be invalid
because of occlusion or because of mismatch.
\begin{equation}
  \textbf{D}_b {\textbf{p}} = \begin{dcases*} 
        \textbf{D}'_{b \textbf{p} }  &   if $ | \textbf{D}'_{b\textbf{p}} - \textbf{D}'_{m \textbf{q}} | \leq 1 $ \\
        invalid & otherwise
        \end{dcases*} 
 \label{eq:lrrl}
\end{equation}

where $\textbf{q} = [ \textbf{p}_x+\textbf{D}'_b\textbf{p},
  \textbf{p}_y]^T $.  The same thing is done for finding the
invalid pixels in the match image simply by reversing the
subscripts $b$ and $m$ in Eq. \ref{eq:lrrl}.


\subsection{More Global Matching (MGM)}
Facciolo et al. \cite{facciolo2015mgm} observed that the
star-shaped support region of SGM was not always sufficient
for the neighboring pixel disparities to sufficiently
constrain each other for the minimization of the
Discontinuity Cost. As a result, one could sometimes see
streaking artifacts in disparity maps.  They got around the
problem by an alternative formulation of the aggregation
step which results in what may be referred to as
``quad-based'' support regions shown in Fig. \ref{fig:MGM}.

The remarkable thing about the aggregation strategy proposed
by Facciolo et al. \cite{facciolo2015mgm} is that it
requires only one more lookup in order to create area-based
support regions of the sort shown in Fig. \ref{fig:MGM}.  This
additional lookup is denoted by the symbol
$\textbf{r}^{\perp}$ in the following energy expression
  that is minimized in MGM:
\begin{align}
\mathcal{L}_{\textbf{r}} (\textbf{p},d) &= \mathcal{C}(\textbf{p},d)+ \frac{1}{2}  \sum_{\textbf{x} \in \textbf{r}, \textbf{r}^{\perp}} \min \Big( \mathcal{L}_{\textbf{x}}(\textbf{p}-\textbf{x},d), \notag \\  & \mathcal{L}_{\textbf{x}}(\textbf{p}-\textbf{x},d-1) +P_1,
 \mathcal{L}_{\textbf{x}}(\textbf{p}-\textbf{x},d+1)+P_1, \notag \\ & \min_{i} \mathcal{L}_{\textbf{x}}(\textbf{p}-\textbf{x},i)+P_2 \Big) 
\label{eq3}
\end{align}
Now instead of just aggregating the costs along a given
direction $\textbf{r}$, we take the average of the value
supplied by the direction $\textbf{r}$ and a direction
$\textbf{r}^{\perp}$ which is perpendicular to
$\textbf{r}$. Fig. \ref{fig:MGM} shows the example of 8
scanning paths in SGM versus those in MGM. Since MGM
recursively aggregates the cost from the original SGM
direction and a direction perpendicular to it, it covers an
area as shown for each direction. Note that for each SGM
direction a perpendicular along the anti-clockwise direction
is selected to produce the quad-area based
coverage. Algorithm \ref{algo:MGM} summarizes the steps of
MGM.
\begin{algorithm}
    \caption{MGM Algorithm} \label{algo:MGM}
 \hspace*{\algorithmicindent} \textbf{Input:} $\textbf{I}_b$,$\textbf{I}_m$, $d_{min}$, $d_{max}$, $\textbf{r}$,$\textbf{r}^{\perp}$ \\
 \hspace*{\algorithmicindent} \textbf{Output:} $\textbf{D}_b$,$\textbf{D}_m$
{\small  \begin{algorithmic}[1]
 \Procedure{MGM}{}
\State $\mathcal{C} = census\_cost(\textbf{I}_b,\textbf{I}_m, d_{min}, d_{max})$
\State $canny = canny\_filter(\textbf{I}_b)$
\State $\mathcal{S}= aggregate\_cost(\mathcal{C}, \textbf{r},\textbf{r}^{\perp},canny )$
\State $\textbf{D}'_b = Compute\_disp(\mathcal{S})$
\State $Median\_filter(\textbf{D}'_b,3)$
\State $\mathcal{C} = census\_cost(\textbf{I}_m,\textbf{I}_b, -d_{max}, -d_{min})$
\State $canny = canny\_filter(\textbf{I}_m)$
\State $\mathcal{S}= aggregate\_cost(\mathcal{C},\textbf{r},\textbf{r}^{\perp}, canny )$
\State $\textbf{D}'_m = Compute\_disp(\mathcal{S})$
\State $Median\_filter(\textbf{D}'_m,3)$
  \EndProcedure
 \end{algorithmic} }
    \end{algorithm}    

\subsection{tSGM and tMGM}
\label{sec:tSGM_tMGM}
A great deal of SGM computation is related to
Eq. (\ref{eq2}) being iterated over all possible disparities
at every pixel.  This is made computationally more efficient
in tSGM by Rothermel \cite{rothermel2012sure} by setting
dynamically calculated bounds for possible disparities in a
hierarchical examination of the image.  At the coarsest
level, Rothermel set the disparity search bound to $[0, W]$
where $W$ is the image width.  Descending down the
hierarchy, the disparity search bound at each pixel at level
$l$ is based on the min and the max of the disparities
estimated in a window of a specified size at level $l-1$.
We are using index 0 for the coarsest level.  This logic is
used for valid pixels only.  As far as invalid pixels are
concerned, for each such pixel at the current level, you
select the median of the disparities at all the valid pixels
in a $31 \times 31$ window at the next coarser level. If we
denote this median value by $m$, the disparity
search bound at the current pixel is set to $[m -
  \tau, m + \tau]$ for a user-specified $\tau$.
Should such a median not be available, you replace $m$ by the average of the disparities calculated over the
entire image at the next coarser level.

Rothermel has applied his implementation of tSGM to aerial
images; the qualitative results obtained by the author are
reported in \cite{rothermel2012sure}.\footnote{To the best
  of our knowledge, there has not yet been a quantitative
  assessment of the performance of this algorithm.}  Our
experience with tSGM based on the implementation described
in \cite{rothermel2012sure} was that it did not generalize
well to the Middlebury dataset we have used in the
comparative results reported in this paper.  In order to
rectify this issue, we have had to make certain changes to
the algorithm itself\footnote{We did try to download the
  executable that is made available by Rothermel
  \cite{SURE_exe}, but it requires camera projection
  matrices that are not available for the Middlebury
  dataset.  In addition, the executable produces the final
  point cloud that is a result of stereo fusion.  It is not
  clear how to change the behavior of the executable in
  order to produce the disparity maps.}  --- while
preserving the hierarchical calculation of the disparity
maps.  We made two key changes: (1) initialization of the
disparity bounds; and (2) propagation of the disparity
bounds through the hierarchy.  Our initialization logic for
the disparity search bounds is described in Table
\ref{tbl:tSGM} and our strategy to propagate the disparity
bounds is explained below.

For valid pixels, we have modified the disparity bound value
at the current pixel at a given level by enforcing the
constraint that both the upper and the lower bounds do not
violate a user-specified global constraint, which for public
datasets are supplied by the dataset creators.  More
specifically, using the estimated disparity map at the
previous level in the resolution hierarchy, the disparity
range at the current level is estimated at each valid pixel
using the local min and max disparity values in a $7 \times
7$ window around the pixel, denoted as
$w_{\textbf{p}_{min}}^L$ and $w_{\textbf{p}_{max}}^L$,
respectively. Then $T^{ min}_{\textbf{p}}$ and $T^{
  max}_{\textbf{p}}$ are updated as follows:

\begin{align}
  \begin{split}
T^{ min}_{\textbf{p}} &= max(w_{\textbf{p}_{min}}^L - \epsilon, \lfloor d_{min}/s \rfloor - \epsilon) \\
T^{ max}_{\textbf{p}} &= min(w_{\textbf{p}_{max}}^L + \epsilon, \lceil d_{max}/s \rceil + \epsilon)
  \end{split}
  \label{eq:tsgm_valid}
\end{align}
where $\epsilon$ is a relaxation parameter and $s$ is the
scale factor at the current level.  The parameters $d_{min}$
and $d_{max}$ are supplied with the dataset images for the
min and the max bounds for disparity search.

For an invalid pixel $\textbf{p}$, the values of $T^{
  min}_{\textbf{p}}$ and $T^{ max}_{\textbf{p}}$ are set to
$ \lfloor d_{min}/s \rfloor - \epsilon$ and $ \lceil
d_{max}/s \rceil + \epsilon$ respectively. This amounts to
searching the complete range $[d_{min}/s, d_{max}/s]$ for
invalid pixels, thereby reducing artifacts and creating
denser disparity maps. 

Setting the disparity search range as indicated above does
affect the overall disparity estimation accuracy. If the
range is set too high (e.g. the image width) then we have a
higher chance of getting false positives, and if it is set
too low, we may lose some important scene details. More
precisely, the data cost term will have greater ambiguities
in a larger disparity search range, since a pixel can be
matched to multiple pixels along the epipolar line. On the
other hand, if it is set too low for a pixel, then the
corresponding matching pixel in the second image may be
outside the search bound -- resulting in a mismatch or
getting marked as an occluded pixel, when it's not occluded.

As we will show in Section \ref{sec:tSGM_tMGM}, implementing the SGM
logic in a hierarchy not only improves the time performance for stereo
matching, but also results in a disparity map that has fewer errors.
That leads to the following questions: If a hierarchical
implementation improves both the time and matching performance of the
basic SGM algorithm, why not do the same with MGM?  Why not create a
new version of MGM that can benefit from the overall speedup that can
be achieved when its logic is implemented in a hierarchy?
Additionally, as we see in tSGM, why not boost the matching
performance of the basic MGM logic by the multi-scale census transform
that would get used in a hierarchical implementation?

Our implementation of tMGM is the answer to these questions.
The new algorithm tMGM initializes the disparity search
bounds in the same manner as in tSGM (ref. Table
\ref{tbl:tSGM}).  Since the matching logic operates in a
hierarchy in tMGM, the matching precision can benefit from
the census transform being applied automatically on a
multi-scale basis.  When this is combined with the power
that tMGM derives from the ``quad-area'' based support
structure, we end up with an algorithm that is superior to
all others. 

The reader may think while the precision in stereo matching
may be expected to go up because of a combination of
``quad-area'' based support and the multi-scale census
transform, the price to pay for that would the computational
time --- on account of the same ``quad-area'' based support
structure.  As we report in Section \ref{sec:Results}, the
time performance of tMGM is only marginally slower than that
of tSGM.  As we will show in that section, comparing just
SGM and MGM, the latter is slower by around 8-13\%.  And
comparing tSGM and tMGM, the latter is slower by around
6-10\%.  To connect these two comparisons, when comparing
SGM with tSGM, the latter is faster by around 50-370\%.

\begin{table}
{\small \begin{supertabular}{p{3.5cm} | p{4.5cm}}
\hline
Initialization & Parameter Definition \\
\hline
$  \textbf{D}_{b{\textbf{p}}} = 0$ & Disparity at position $\textbf{p}$ in base image \\
$\textbf{D}_{m{\textbf{p}}} = 0$ & Disparity at position $\textbf{p}$ in match image \\
$\textbf{T}_{b{\textbf{p}}}^{min} = \lfloor d_{min}/s \rfloor - \epsilon$ &  Lower search bound at pixel $\textbf{p}$ in base image.\\
$ \textbf{T}_{b{\textbf{p}}}^{max} = \lceil d_{max}/s \rceil + \epsilon $ & Upper search bound at pixel $\textbf{p}$ in base image. \\
$ \textbf{T}_{m{\textbf{p}}}^{min} = \lfloor -d_{max}/s \rfloor - \epsilon$ &  Lower search bound at pixel $\textbf{p}$ in match image.\\
$\textbf{T}_{m{\textbf{p}}}^{max} = \lceil -d_{min}/s \rceil + \epsilon$ &  Upper search bound at pixel $\textbf{p}$ in match image.\\
\hline
\end{supertabular} }
\caption{tSGM and tMGM initialization. $s$, $d_{min}$ and
  $d_{max}$ are input parameters to tSGM and tMGM algorithms
  (see Algorithms \ref{algo:tSGM} and \ref{algo:tMGM}) and
  $\epsilon$ is a relaxation parameter.}
\label{tbl:tSGM}
\end{table}

\begin{algorithm}
    \caption{tSGM Algorithm}\label{algo:tSGM}
 \hspace*{\algorithmicindent} \textbf{Input:} $\textbf{I}_b$,$\textbf{I}_m$, $\textbf{T}_{b}^{max}$, $\textbf{T}_{b}^{min}$, $\textbf{T}_{m}^{max}$, $\textbf{T}_{m}^{min}$,$d_{min}$, $d_{max}$, scale factor $s$ (All inputs Downsampled by $s$) \\
 \hspace*{\algorithmicindent} \textbf{Output:} $\textbf{D}_b$, $\textbf{D}_m$

{\small  \begin{algorithmic}[1]
 \Procedure{tSGM iterations}{}
\While{$s \neq 0$}
\State $\textbf{D}'_b$ = $SGM(\textbf{I}_b,\textbf{I}_m, \textbf{T}_{b}^{max}$, $\textbf{T}_{b}^{min})$ \Comment{For all $\bf{r}$}
\State $\textbf{D}'_m$ = $SGM(\textbf{I}_m,\textbf{I}_b, \textbf{T}_{m}^{max}$, $\textbf{T}_{m}^{min})$ \Comment{For all $\bf{r}$}
\State $ \textbf{D}_b = LRRL(\textbf{D}'_b,\textbf{D}'_m)$
\State $ \textbf{D}_m = LRRL(\textbf{D}'_m,\textbf{D}'_b)$

\State $s = \lfloor s / 2 \rfloor$ 
\State $\textbf{D}_b = 2* \textbf{D}_b$
\State $\textbf{D}_m = 2* \textbf{D}_m$
\State $Resize_s(\textbf{I}_b, \textbf{I}_m,\textbf{T}_{b}^{min},\textbf{T}_{b}^{max},\textbf{T}_{m}^{min},\textbf{T}_{m}^{max})$ \Comment Upscale by 2
\State $\textbf{T}_{b}^{max}$, $\textbf{T}_{b}^{min}= Update(\textbf{D}_b, d_{min}, d_{max})$ \Comment{Update limits }
\State $\textbf{T}_{m}^{max}$, $\textbf{T}_{m}^{min}= Update(\textbf{D}_m, -d_{max}, -d_{min})$ \Comment{Update limits }
\EndWhile
  \EndProcedure
 \end{algorithmic} }
    \end{algorithm}

\begin{algorithm}
    \caption{tMGM Algorithm}\label{algo:tMGM}
 \hspace*{\algorithmicindent} \textbf{Input:} $\textbf{I}_b$,$\textbf{I}_m$, $\textbf{T}_{b}^{max}$, $\textbf{T}_{b}^{min}$, $\textbf{T}_{m}^{max}$, $\textbf{T}_{m}^{min}$,$d_{min}$, $d_{max}$, scale factor $s$ (All inputs Downsampled by $s$) \\
 \hspace*{\algorithmicindent} \textbf{Output:} $\textbf{D}_b$, $\textbf{D}_m$

{\small  \begin{algorithmic}[1]
 \Procedure{tMGM iterations}{}
\While{$s \neq 0$}
\State $\textbf{D}'_b$ = $MGM(\textbf{I}_b,\textbf{I}_m, \textbf{T}_{b}^{max}$, $\textbf{T}_{b}^{min})$ \Comment{For all $\bf{r}$ and $\bf{r}^\perp$}
\State $\textbf{D}'_m$ = $MGM(\textbf{I}_m,\textbf{I}_b, \textbf{T}_{m}^{max}$, $\textbf{T}_{m}^{min})$ \Comment{For all $\bf{r}$ and $\bf{r}^\perp$}
\State $ \textbf{D}_b = LRRL(\textbf{D}'_b,\textbf{D}'_m)$
\State $ \textbf{D}_m = LRRL(\textbf{D}'_m,\textbf{D}'_b)$

\State $s = \lfloor s / 2 \rfloor$ 
\State $\textbf{D}_b = 2* \textbf{D}_b$
\State $\textbf{D}_m = 2* \textbf{D}_m$
\State $Resize_s(\textbf{I}_b, \textbf{I}_m,\textbf{T}_{b}^{min},\textbf{T}_{b}^{max},\textbf{T}_{m}^{min},\textbf{T}_{m}^{max})$ \Comment Upscale by 2
\State $\textbf{T}_{b}^{max}$, $\textbf{T}_{b}^{min}= Update(\textbf{D}_b, d_{min}, d_{max})$ \Comment{Update limits }
\State $\textbf{T}_{m}^{max}$, $\textbf{T}_{m}^{min}= Update(\textbf{D}_m, -d_{max}, -d_{min})$ \Comment{Update limits }
\EndWhile
  \EndProcedure
 \end{algorithmic} }
    \end{algorithm}
\begin{figure}
\begin{center}
\includegraphics[width=\linewidth]{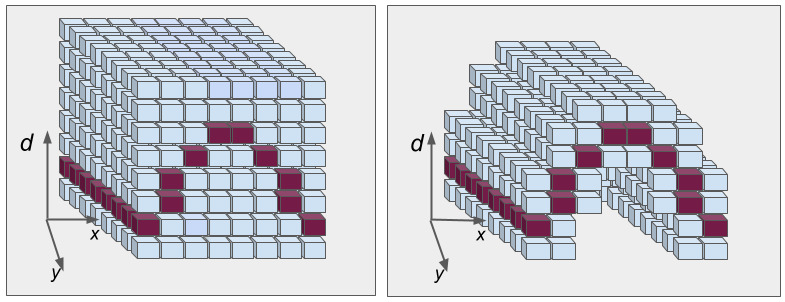}
\end{center}
\caption{Search volume in SGM and MGM vis-a-vis that in tSGM
  and tMGM. Whereas SGM and MGM use images at their original
  resolution, tSGM and tMGM use image resolution hierarchies
  in which each coarse level is used to estimate the
  disparity bounds to be used for the search in the next
  finer level.}
\label{fig:tSGM}
\end{figure}
\begin{table*} 
\begin{tabular}{p{1.8cm}|p{2.5cm}|p{1.0cm}| p{0.8cm}||p{0.8cm}p{0.8cm}|p{0.8cm}p{0.8cm}|p{0.8cm}p{0.8cm}}
\toprule
\multicolumn{1}{c}{} & \multicolumn{1}{|c}{} & \multicolumn{1}{|c}{} & \multicolumn{1}{|c||}{} & \multicolumn{2}{c}{$>1$ pixel error} & \multicolumn{2}{|c}{$>2$ pixel error} & \multicolumn{2}{|c}{$>3$ pixel error} \\
\cline{5-10} 
Algo  & Img & Avg Disp Err (pix)  & Inv Pix Err (\%) &  Bad Pix Err (\%) &  Tot Err (\%)  & Bad Pix Err (\%) &  Tot Err (\%)  & Bad Pix Err (\%) &  Tot Err (\%)   \\ \hline
\multirow{3}{1.4cm}{SGM-8} & MB2014 (P) & 7.71 & 38.11 & \textbf{14.19} & 52.30 & \textbf{8.05} & 46.16 & \textbf{6.72} & 44.83  \\ 
 & MB2014 (P\_E) & 10.11 & 41.48 & \textbf{15.31} & 56.78 & \textbf{8.17} & 49.65 & \textbf{6.82} & 48.29  \\ 
 & MB2014 (P\_L) & 17.57 & 55.66 & \textbf{16.29} & 71.95 & \textbf{10.29} & 65.95 & \textbf{9.00} & 64.67  \\ 
 & KITTI2015 & 1.64 & 19.90 & \textbf{24.97} & 44.87 & \textbf{8.96} & 28.87 & \textbf{5.07} & 24.98  \\ 
 & ETH3D & 2.15 & 19.61 & \textbf{6.02} & 25.64 & 3.02 & 22.63 & 2.47 & 22.08  \\ 
\hline 
\multirow{3}{1.4cm}{MGM-8} & MB2014 (P) & 9.75 & 35.42 & 15.17 & 50.60 & 9.15 & 44.57 & 7.88 & 43.30  \\ 
 & MB2014 (P\_E) & 13.06 & 38.93 & 16.48 & 55.41 & 9.51 & 48.45 & 8.21 & 47.15  \\ 
 & MB2014 (P\_L) & 22.56 & 52.72 & 18.36 & 71.09 & 12.51 & 65.24 & 11.27 & 64.00  \\ 
 & KITTI2015 & 1.85 & 18.17 & 29.21 & 47.38 & 10.06 & 28.23 & 5.27 & 23.44  \\ 
 & ETH3D & 3.24 & 18.36 & 6.39 & 24.74 & 3.82 & 22.17 & 3.28 & 21.64  \\ 
\hline 
\multirow{4}{1.4cm}{SGM-16} & MB2014 (P) & 6.71 & 32.42 & 16.53 & 48.95 & 9.02 & 41.44 & 7.28 & 39.70  \\ 
 & MB2014 (P\_E) & 8.91 & 35.95 & 17.77 & 53.72 & 9.31 & 45.27 & 7.58 & 43.53  \\ 
 & MB2014 (P\_L) & 15.34 & 49.32 & 18.80 & 68.13 & 11.43 & 60.76 & 9.71 & 59.03  \\ 
 & KITTI2015 & 1.75 & 19.62 & 31.04 & 50.66 & 13.25 & 32.87 & 7.50 & 27.12  \\ 
 & ETH3D & 1.30 & 14.42 & 6.61 & 21.04 & \textbf{2.99} & 17.42 & \textbf{2.30} & 16.73  \\ 
\hline 
\multirow{3}{1.4cm}{MGM-16} & MB2014 (P) & 6.93 & 29.92 & 15.90 & 45.82 & 8.78 & 38.71 & 7.22 & 37.15  \\ 
 & MB2014 (P\_E) & 9.39 & 33.72 & 17.54 & 51.26 & 9.33 & 43.06 & 7.73 & 41.45  \\ 
 & MB2014 (P\_L) & 17.09 & 46.52 & 19.38 & 65.90 & 12.18 & 58.70 & 10.58 & 57.10  \\ 
 & KITTI2015 & 1.66 & 15.10 & 32.26 & 47.36 & 11.76 & 26.86 & 6.05 & 21.15  \\ 
 & ETH3D & 1.71 & 13.80 & 6.24 & 20.04 & 3.24 & 17.04 & 2.66 & 16.46  \\ 
\hline 
\multirow{3}{1.4cm}{tSGM-8} & MB2014 (P) & \textbf{5.58} & 23.28 & 21.15 & 44.44 & 12.01 & 35.29 & 9.37 & 32.65  \\ 
 & MB2014 (P\_E) & \textbf{6.47} & 25.37 & 23.27 & 48.64 & 12.68 & 38.05 & 9.87 & 35.24  \\ 
 & MB2014 (P\_L) & \textbf{15.28} & 36.52 & 27.06 & 63.58 & 17.66 & 54.17 & 14.85 & 51.37  \\ 
 & KITTI2015 & \textbf{1.27} & 12.53 & 27.63 & \textbf{40.16} & 10.14 & \textbf{22.67} & 5.81 & 18.34  \\ 
 & ETH3D & 0.76 & 8.17 & 9.93 & 18.10 & 4.34 & 12.50 & 2.65 & 10.82  \\ 
\hline 
\multirow{3}{1.4cm}{tMGM-8} & MB2014 (P) & 5.59 & 21.52 & 21.17 & 42.70 & 12.18 & 33.70 & 9.57 & 31.10  \\ 
 & MB2014 (P\_E) & 6.59 & 23.48 & 23.38 & 46.86 & 12.97 & 36.45 & 10.19 & 33.67  \\ 
 & MB2014 (P\_L) & 16.71 & 35.23 & 27.43 & 62.65 & 18.27 & 53.50 & 15.54 & 50.77  \\ 
 & KITTI2015 & 1.34 & 12.36 & 31.95 & 44.31 & 11.27 & 23.63 & 5.91 & \textbf{18.27}  \\ 
 & ETH3D & 0.84 & 7.65 & 9.95 & 17.60 & 4.99 & 12.64 & 3.16 & 10.81  \\ 
\hline 
\multirow{4}{1.4cm}{tSGM-16} & MB2014 (P) & 6.32 & 20.58 & 23.41 & 43.99 & 13.86 & 34.44 & 11.11 & 31.70  \\ 
 & MB2014 (P\_E) & 7.32 & 22.26 & 25.74 & 48.00 & 14.80 & 37.06 & 11.92 & 34.18  \\ 
 & MB2014 (P\_L) & 15.68 & 31.40 & 30.52 & 61.91 & 20.36 & 51.76 & 17.22 & 48.62  \\ 
 & KITTI2015 & 1.71 & 15.22 & 34.70 & 49.92 & 16.15 & 31.37 & 10.05 & 25.27  \\ 
 & ETH3D & \textbf{0.72} & 6.27 & 9.37 & 15.64 & 3.86 & 10.13 & 2.46 & 8.73  \\ 
\hline 
\multirow{3}{1.4cm}{tMGM-16} & MB2014 (P) & 5.87 & \textbf{19.08} & 22.44 & \textbf{41.53} & 13.15 & \textbf{32.23} & 10.49 & \textbf{29.57}  \\ 
 & MB2014 (P\_E) & 6.81 & \textbf{20.59} & 24.82 & \textbf{45.42} & 14.07 & \textbf{34.66} & 11.25 & \textbf{31.85}  \\ 
 & MB2014 (P\_L) & 15.80 & \textbf{30.36} & 29.78 & \textbf{60.14} & 19.80 & \textbf{50.15} & 16.79 & \textbf{47.14}  \\ 
 & KITTI2015 & 1.47 & \textbf{11.79} & 34.22 & 46.01 & 13.02 & 24.81 & 7.11 & 18.89  \\ 
 & ETH3D & 0.73 & \textbf{5.77} & 8.92 & \textbf{14.69} & 4.08 & \textbf{9.85} & 2.65 & \textbf{8.42}  \\ 
\hline 
 
\bottomrule
\end{tabular} 

\caption{Errors averaged over all the image pairs in each
  dataset. MB stands for Middlebury. For the Middlebury 2014
  dataset, this table only considers the image pairs with
  perfect rectification and with groundtruth. (There are 23
  such image pairs out of a total of 33.)  The notation
  "P\_E" stands for the image pairs with exposure variation
  in views, and the notation "P\_L" is for the image pairs
  with illumination variations. Note that bad pixel errors
  only consider pixels that are valid in both
  $\textbf{D}_{gt}$ and $\textbf{D}_b$.  Note that for the
  Kitti2015 and ETH3D we used only the training images.}

\label{tbl:avg_err}
\end{table*} 
%

\setlength{\aboverulesep}{0pt}
\setlength{\belowrulesep}{0pt}

\setlength{\aboverulesep}{0pt}
\setlength{\belowrulesep}{0pt}

Fig. \ref{fig:tSGM} illustrates a comparison between the
search space in SGM and MGM vis-a-vis the search space for
tSGM and tMGM. For SGM and MGM, the search space is a full
cuboidal volume whose width and height are the same as for
the images and whose depth equals the search bound for the
disparities.  On the other hand, for tSGM and tMGM, the
width and the height remain the same except for the effect
of downsampling at the coarser levels of the hierarchy, and
the depth varies from pixel to pixel since, at each level,
it is set according to the disparity map at the next coarser
level.  At the coarsest level, the depth is constant and set
according to Table \ref{tbl:tSGM}.


\section{Comparative Evaluation}

This section is organized as follows. We first describe the
datasets that we use for our exhaustive evaluation.
We then present the results of a comparative evaluation of
the different algorithms on the Middlebury2014
\cite{scharstein2014high}, KITTI2015 \cite{Menze2015ISA} and
ETH3D {\cite{schoeps2017cvpr}} two-view stereo datasets
using the Middlebury V3 benchmark. This is followed by a
separate subsection on our experiments using satellite
images from the MVS Challenge dataset \cite{MVS_Challenge}.

\subsection{Datasets and Implementation}

For the Middlebury2014 dataset, we use the 23 full
resolution (5-6 MP) image pairs for which the groundtruth is
available. These images mainly feature indoor scenes. For
each image pair, there are six ways of pairing the two
images: (1) with perfect rectification; (2) with imperfect
rectification; (3) perfect rectification with differences in
illumination; (4) imperfect rectification with differences in
illumination; (5) perfect rectification with exposure
differences; and, finally, (6) imperfect rectification with
exposure differences.  Imperfect rectification means that
the rows are not aligned correctly due to, say, camera
calibration errors.

The KITTI2015 dataset consists of 200 image pairs of outdoor
scenes for which we have the groundtruth.  This dataset
provides semi-dense groundtruth disparity maps over roughly
30\% of all the pixels.  The pixels where the groundtruth is
available are generally in the bottom half of the images on
account of the fact that images are generated with vehicle
mounted cameras.

The ETH3D dataset consists of 27 image pairs with dense
ground truth disparity maps, featuring both indoor and
outdoor scenes. All 27 image pairs that we have used are
from the training dataset for which the groundtruth is
available.

The comparative results we show in this section are all
based on our own implementations of all four matching
algorithms.  Having our own implementation enables us to use
consistent code for the cross-comparison evaluation and to
easily experiment with the different parameters that specify
the shape of the support structure. For sanity check, we
have compared the results we obtain with our implementation
with those produced by the code provided by Facciolo for
SGM-8 and MGM-8 \cite{MGM_code}. For SGM-8 and MGM-8, we
obtain similar results in both cases.

We use the overcounting correction proposed by Drory et
al. \cite{drory2014semi} in all the algorithms in our
evaluation. Referring to Eq. \ref{eq:sum_eq} in the baseline
SGM algorithm, the data cost is overcounted in the sum volume
$\mathcal{S}$ as it is a part of each
$\mathcal{L}_{\textbf{r}}$. The correction consists of
updating the volume $\mathcal{S}$ as follows 
$$\mathcal{S}(\textbf{p},d) = \mathcal{S}(\textbf{p},d) -
(N-1)\mathcal{C}(\textbf{p},d)$$
where $N$ is the number of directions, 8 or 16 in our
case.

All algorithms use the Hamming distance based on the census
transform over $5 \times 5$ neighborhoods after it is
normalized by the number of image color channels.  For all
implementations of SGM and MGM, a median filter of size $3
\times 3$ is applied to $\textbf{D}'_b$ and
$\textbf{D}'_m$. For tSGM and tMGM, the median filter is
applied at each level.

Referring to Section 3, we notice that these algorithms have
additional parameters that need to be chosen with some care
and intuition, particularly $P_1$ and $P_2$. We set $P_1 =
24$ and use the Canny filter response to adaptively update
$P_2$.  When the filter response is 1, we use $P_2 = 27$ and
when there is no edge, we use $P_2 = 96$. These values were
chosen empirically to get best results for the baseline
SGM-N algorithm. Additionally, for tSGM and tMGM, we set
$s=8$ and $\epsilon = 4$ where $s$ is the scale factor
defined in Algorithms \ref{algo:tSGM} and \ref{algo:tMGM}
and $\epsilon$ is the relaxation parameter used in
Eq. (\ref{eq:tsgm_valid}).

\subsection{Results}
\label{sec:Results}

In this section, we will first present the metrics we have
used for a quantitative assessment of the disparities as
calculated by the different SGM variants. That will be
followed by a presentation of the quantitative results
obtained with the metrics; these will be conveyed through
the numbers shown in Table \ref{tbl:avg_err}.

Since gross numbers never tell the whole story about how an
algorithm may have performed with respect to the scene and
illumination conditions, the subsections that follow take up
each of those conditions and show the artifacts and the
errors given rise to by those conditions.


For quantitative evaluation, we use the following metrics:
{\em invalid pixel error}, {\em bad pixel error}, {\em total
  error}, and {\em average error} as defined in the
Middlebury V3 Benchmark \cite{MB_V3}.  The invalid
pixel error is defined as the percentage of the valid pixels
in the ground truth disparity map $\textbf{D}_{gt}$, that
are marked as invalid in the estimated disparity map
$\textbf{D}_b$. The bad pixel error is defined as the
percentage of the valid pixels in $\textbf{D}_{gt}$ such
that $|\textbf{D}_{b\textbf{p}} - \textbf{D}_{gt\textbf{p}}|
> \delta$ for a given threshold $\delta$.  Note that a pixel
that is marked invalid in $\textbf{D}_b$ but valid in
$\textbf{D}_{gt}$ is not counted in the bad pixel error
because it has already been counted in the invalid pixel
error. The total error is simply the sum of the invalid
pixel error and the bad pixel error. The average error is
defined as $ \frac{1}{|V|} \sum_V |\textbf{D}_{b\textbf{p}}
- \textbf{D}_{gt\textbf{p}}| $ where $V$ is the set of
pixels that are valid in both $\textbf{D}_{b}$ and
$\textbf{D}_{gt}$.

Table \ref{tbl:avg_err} reports the errors defined as above,
for $\delta \in \{1,2,3\}$ pixels, for the Middlebury2014
dataset with perfect rectification (with and without
exposure or illumination variations), the KITTI2015 and ETH3D
datasets. The imperfect rectification case for the
Middlebury2014 dataset is discussed separately in Subsection
\ref{sec:Imp_rect}.

With regard to the invalid pixel error, as shown in Table
\ref{tbl:avg_err}, tMGM-16 produces the best results in all
cases.  Note that although SGM-8 produces a smaller
bad-pixel-error than the other algorithms, this error does
not account for invalid pixels. With regard to the total
error, which does account for the invalid pixels, tMGM-16 is
superior for all cases except for the KITTI2015 dataset. The
KITTI2015 dataset differs from the other datasets in certain
aspects: the groundtruth disparity maps are semi-dense and
the images are smaller and contain only outdoor
scenes. Another possible reason could be that the values we
chose for P1 and P2 parameters using the Middlebury2014
dataset. It could very well be that tuning P1 and P2 to the
KITTI2015 dataset will bridge the gap between tMGM-16 and
tSGM-8 for that dataset.

Overall, tMGM-16 does indeed benefit from the combined use
of the hierarchical approach, ``quad-area'' based support
structures, and 16 directions along which the costs as given
by Eq. \ref{eq3} are aggregated.  For all the datasets, the
hierarchical versions tSGM-N and tMGM-N achieve lower total
error when compared to their non-hierarchical
counterparts. Interestingly, tSGM-8 produces the lowest
average disparity errors in all cases except for the ETH3D
dataset for which it performs comparably to the best. The
average errors for tMGM-16 are quite close to tSGM-8. 

\begin{table}[h]
\centering
\begin{tabular}{p{1.7cm} | p{1.2cm}p{1.2cm}p{1.2cm}p{1.4cm}}
\hline
Dataset & SGM-8  & MGM-8  & tSGM-8  & tMGM-8\\
\hline
Perfect (M) & 104.1 & 113.3 & 69.9 & 74.5 \\
KITTI2015 & 7.5 & 8.6 & 4.0 & 4.3 \\
ETH3D & 16.4 & 18.3 & 4.4 & 4.9 \\
\hline
\end{tabular}
\caption{Average runtime in seconds}
\label{tbl:tSGM_runtime}
\end{table} 
Table \ref{tbl:tSGM_runtime} presents the run time for each
algorithm. The time shown is averaged over 10 image pairs
for each dataset. As expected, MGM-8 is slower than SGM-8,
and, due to its hierarchical nature, tSGM-8 is faster than
SGM-8. Correspondingly, tMGM-8 is slightly slower than
tSGM-8. Similar observations can be expected for the xGM-16
variants. For all algorithms mentioned in this paper, we
have C++ based implementations with OpenMP support for CPU
parallelization. The runtimes we report are recorded on a
machine with 60GB RAM, and the Intel Broadwell-IBRS @ 2.60
GHz processor with 13 physical cores.

In the subsections that follow, we take up each of the
confounding conditions and show the artifacts that are
caused by each condition for the different variants of SGM.
As mentioned earlier, these confounding conditions are:

illumination variations between the
two views, scenes with untextured or weakly textured regions
and repetitive patterns in the presence of imperfect
rectification.  We illustrate each case with examples.

\begin{figure}[ht]
\centering
    \begin{subfigure}[b]{0.12\textwidth}
        \includegraphics[width=\textwidth]{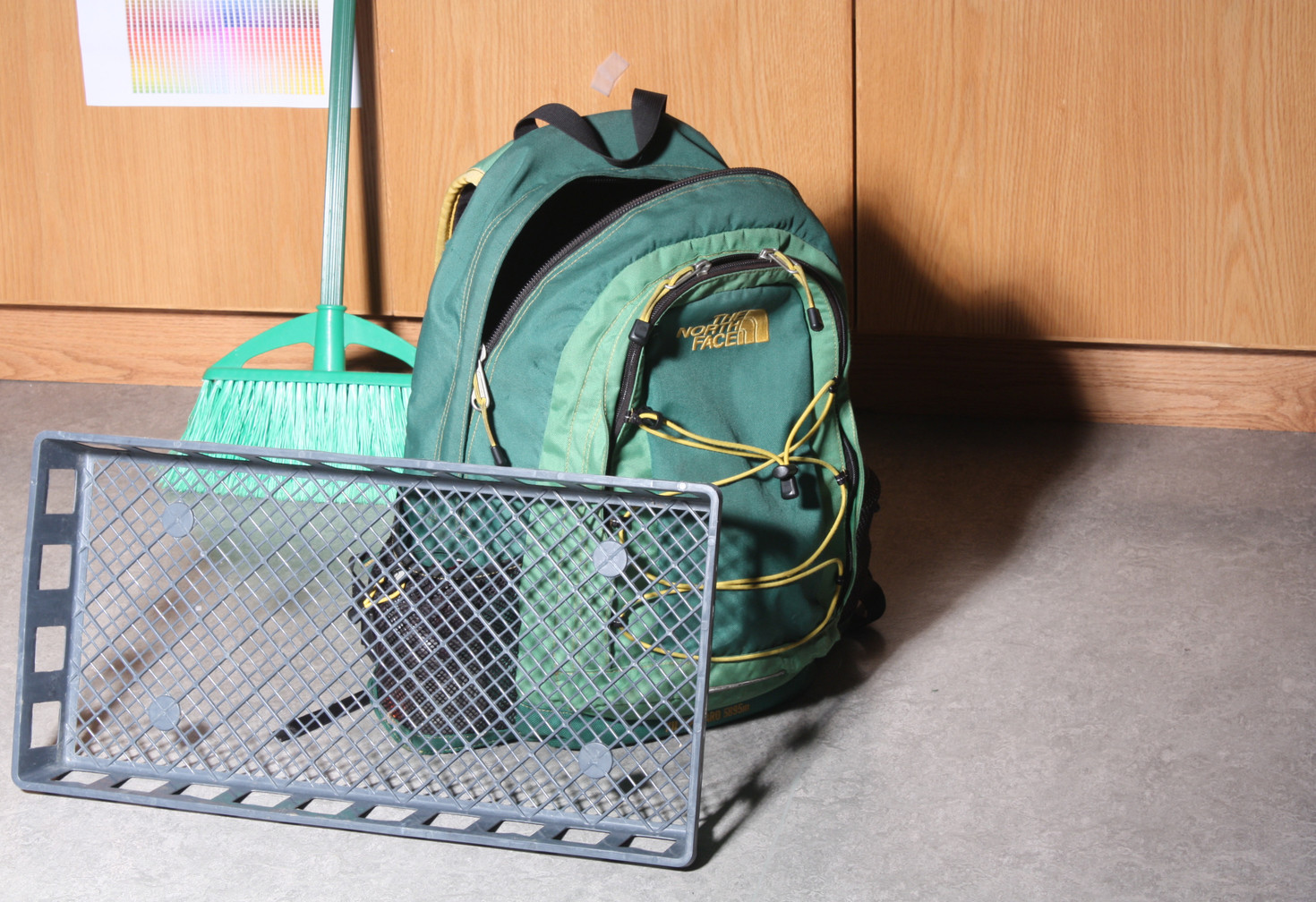}
        \caption{Right Image}
        \label{fig:light_im1}
    \end{subfigure}
    ~ 
    \begin{subfigure}[b]{0.12\textwidth}
        \includegraphics[width=\textwidth]{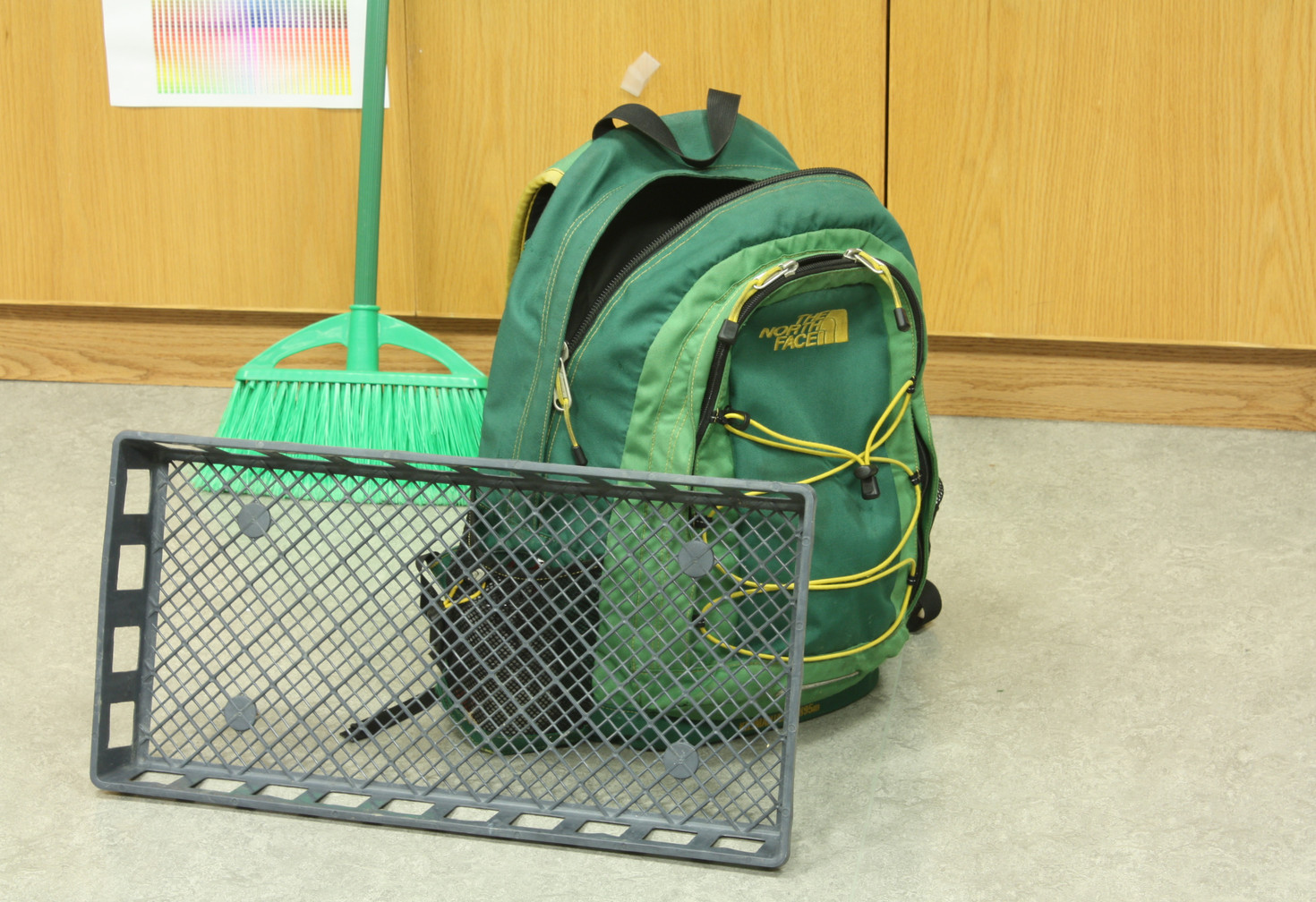}
        \caption{Left Image}
        \label{fig:light_im0}
    \end{subfigure}
    ~
        \begin{subfigure}[b]{0.12\textwidth}
        \includegraphics[width=\textwidth]{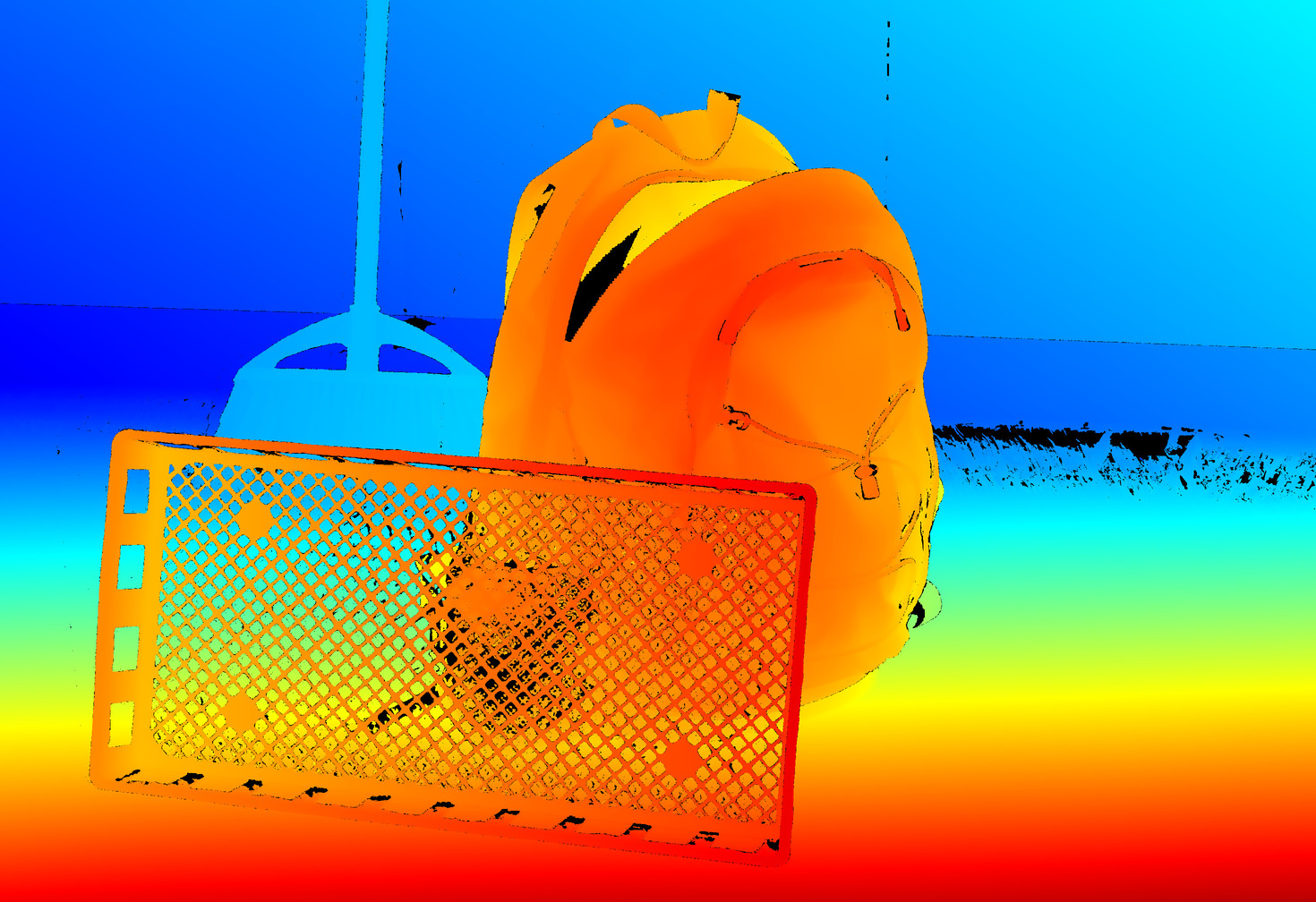}
        \caption{GT Disparity}
        \label{fig:light_disp}
    \end{subfigure}
    \caption{An example from the Middlebury2014 dataset for
      scenes with illumination variations in two views. See
      Figs. \ref{fig:middle_ill_results8} and
      \ref{fig:middle_ill_results16} for the xGM-8 and
      xGM-16 results.}
\label{fig:middle_ill}
\end{figure}

\begin{figure*}
\centering
\includegraphics[width=0.8\textwidth]{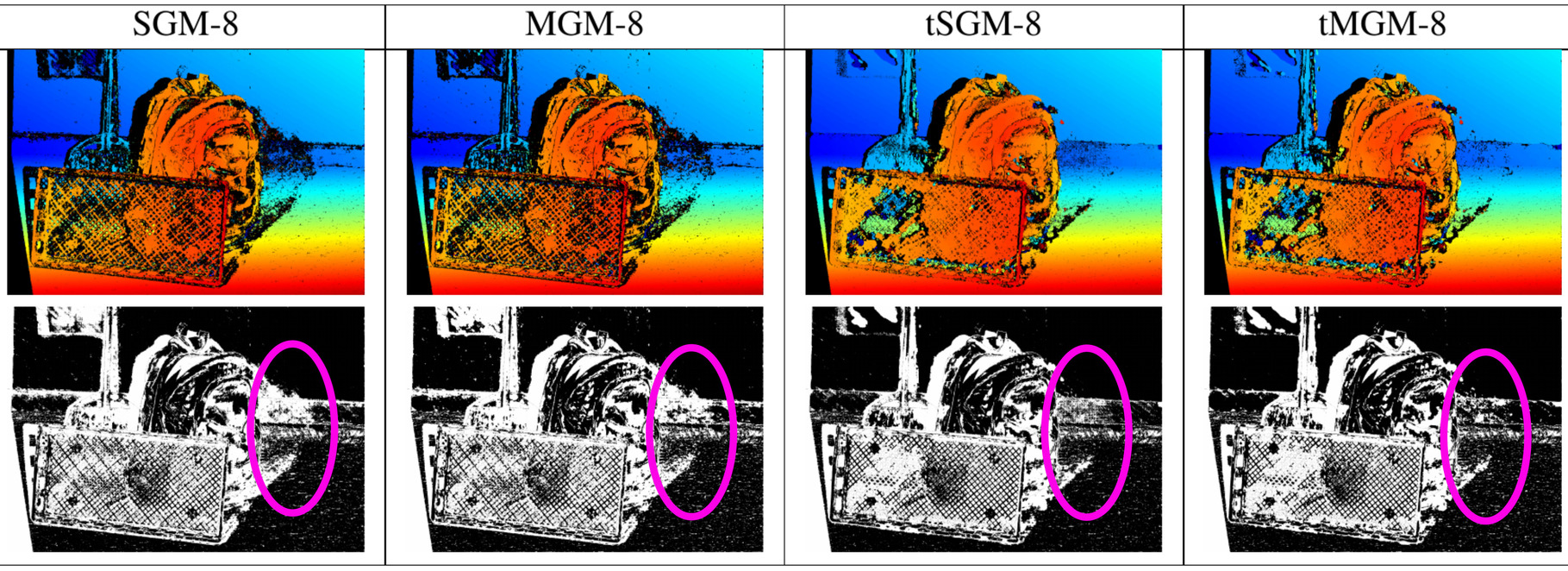}
\caption{Middlebury2014 results using xGM-8 in the presence
  of illumination differences. A region with strong shadows
  is highlighted in magenta. The top row shows the estimated
  disparity maps and the bottom row shows the corresponding
  binary error masks, where a pixel is set to foreground if
  the disparity error $(>1)$ or if it is invalid in
  $\textbf{D}_b$ and valid in $\textbf{D}_{gt}$.}
\label{fig:middle_ill_results8}
\end{figure*}
\begin{figure*}
\centering
\includegraphics[width=0.8\textwidth]{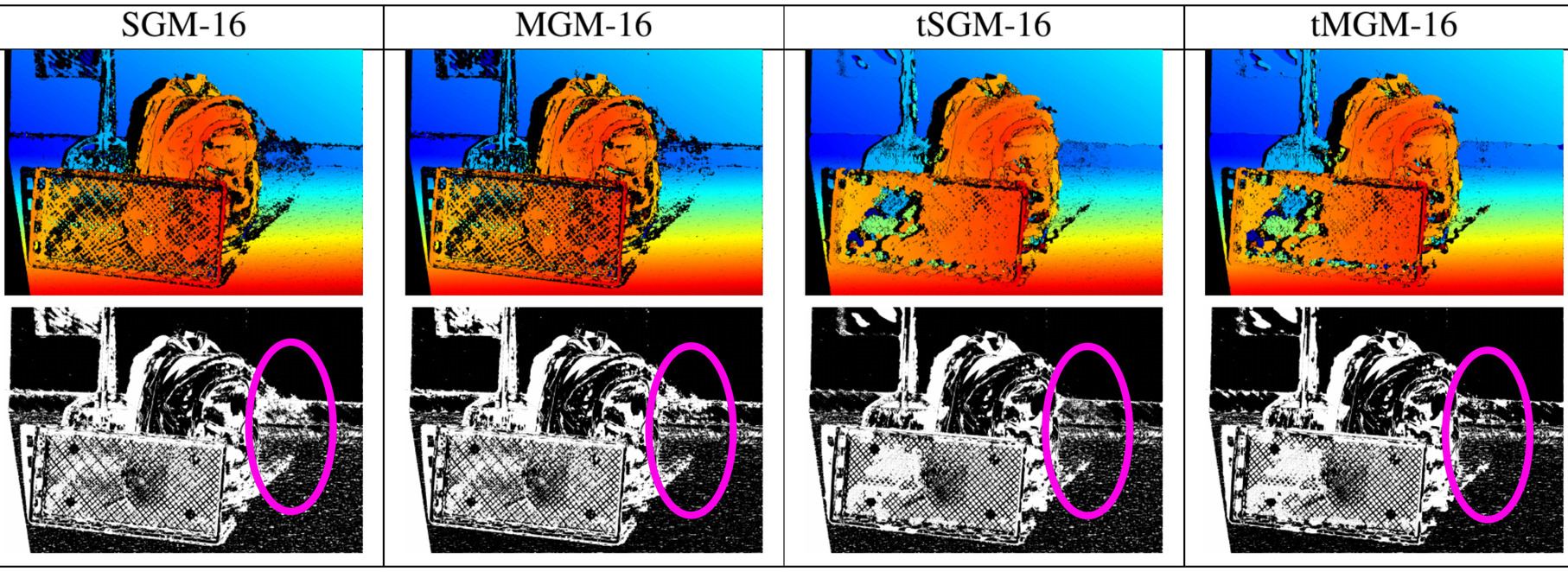}
\caption{Middlebury2014 results using xGM-16 in the presence
  of illumination differences. A region with strong shadows  
is highlighted in magenta. The top row shows the estimated
disparity maps and the bottom row shows the corresponding
binary error masks, where a pixel is
set to foreground if the disparity error $(>1)$ or if it is
invalid in $\textbf{D}_b$ and valid in $\textbf{D}_{gt}$.}
\label{fig:middle_ill_results16}
\end{figure*}

\subsubsection{Illumination Variations \protect \footnote{Illumination variation issues related to out-of-date satellite images are discussed in Section \ref{subsec:Sat} } }

In Table \ref{tbl:avg_err}, ``MB2014 (P\_L)'' represents the
image pairs with perfect rectification in the Middlebury2014
dataset that exhibit illumination differences caused by
strong shadows and by the presence of specularly reflecting
surfaces.  As mentioned earlier in Section \ref{sec:Results},
tMGM-16 produces the lowest total error for this case. In
fact, we observe that all hierarchical variants,
namely tSGM-N and tMGM-N benefit from using the multi-scale
census data cost when compared to their corresponding
non-hierarchical counterparts.  We support this observation
with comparisons in Figs. \ref{fig:middle_ill},
\ref{fig:middle_ill_results8} and
\ref{fig:middle_ill_results16}. In Fig \ref{fig:middle_ill},
we show an example image pair with illumination variation
and the corresponding ground truth disparity map. The
estimated disparity maps and error masks are shown in
Figs. \ref{fig:middle_ill_results8} and
\ref{fig:middle_ill_results16}. Upon inspecting the
highlighted region that contains strong shadows, we see that
the errors are sparser for tMGM-16 which also benefits from
the ``quad-based'' support structure.
\begin{figure}[ht]
\centering
    \begin{subfigure}[b]{0.2\textwidth}
        \includegraphics[width=\textwidth]{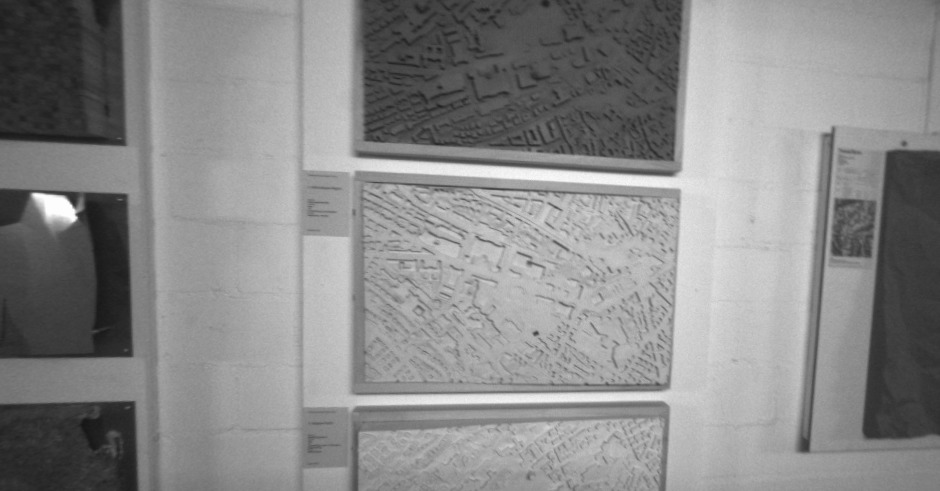}
        \caption{Left Image}
        \label{fig:untextured_im0}
    \end{subfigure}
    ~ 
    \begin{subfigure}[b]{0.2\textwidth}
        \includegraphics[width=\textwidth]{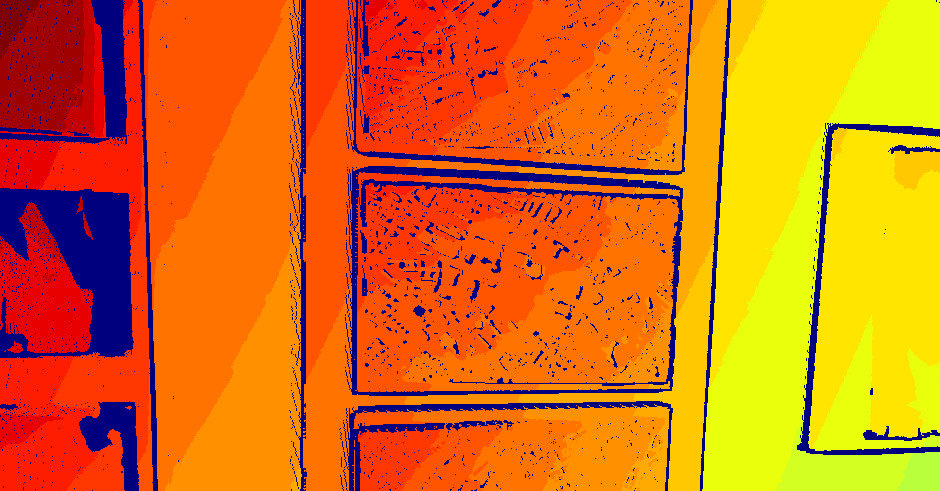}
        \caption{Ground Truth Disparity}
        \label{fig:untextured_gt}
    \end{subfigure}
    \caption{An example for scenes with weakly textured
      regions from the ETH3D dataset }
    \label{fig:ETH_untextured_pair}
\end{figure}


\begin{figure}[h]
\centering
            \begin{subfigure}[b] {0.18\textwidth}
         \includegraphics[width=\textwidth]{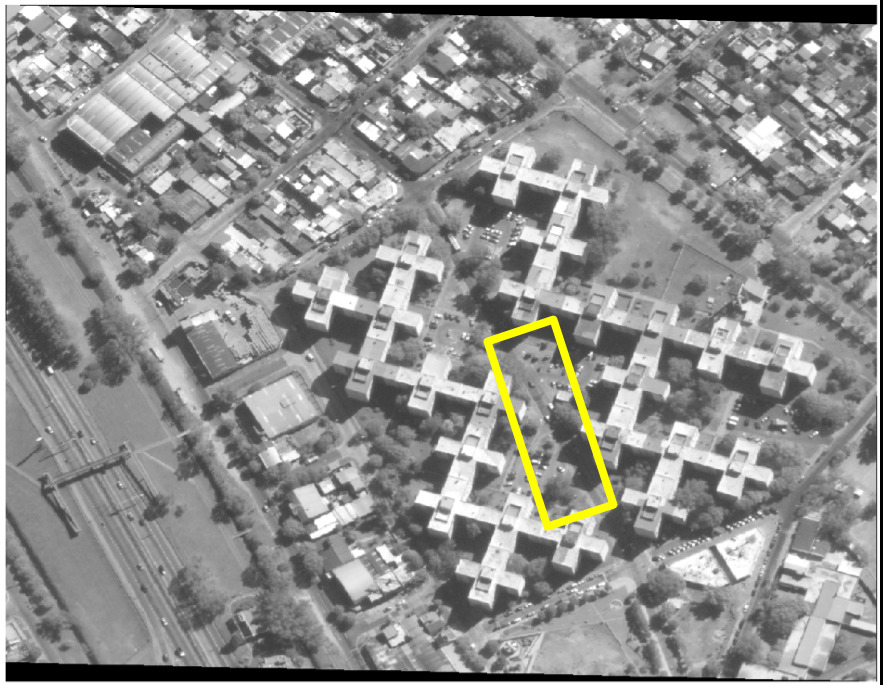} 
         \caption{Unrectified Left Image, Date: 21 MAR 2015 Time:13:57 }    
    \end{subfigure}
    ~          
        \begin{subfigure}[b] {0.19\textwidth}
         \includegraphics[width=\textwidth]{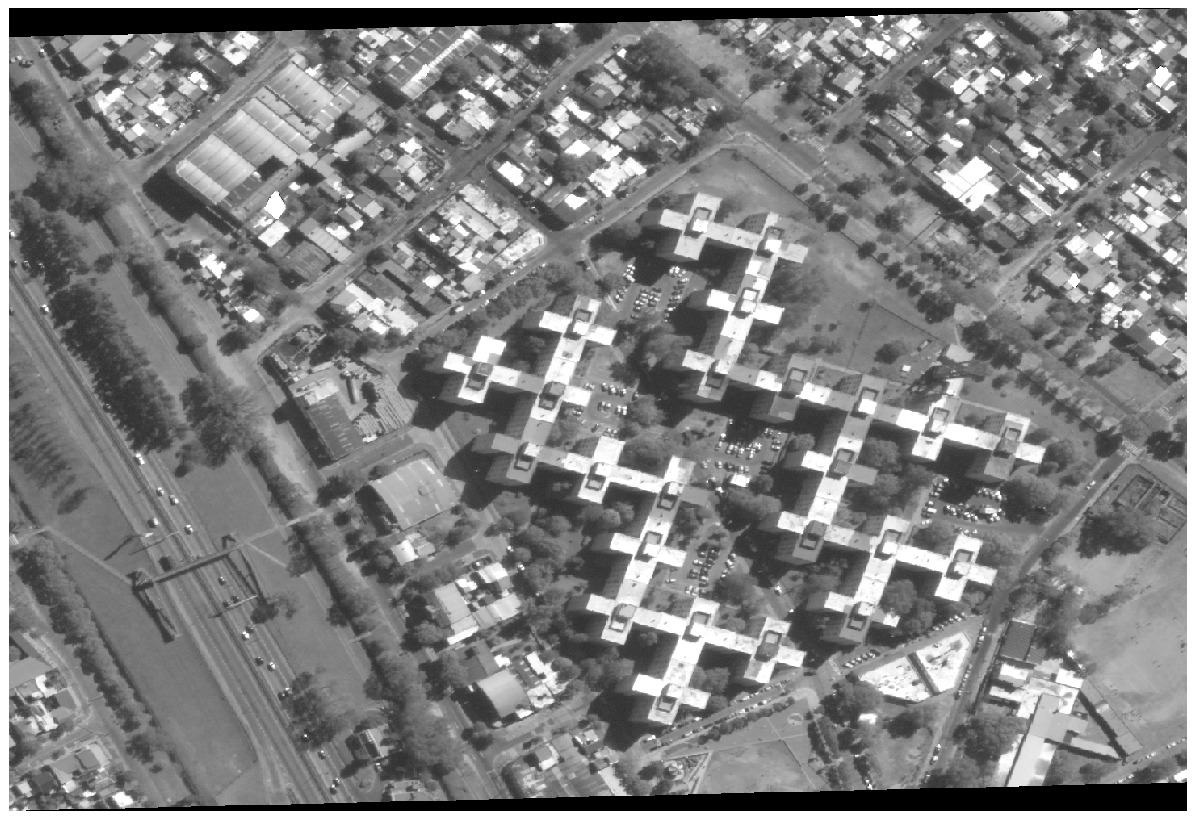} 
         \caption{Unrectified Right Image, Date: 22 MAR 2015, Time: 14:12}    
    \end{subfigure}
    \caption{An example of an out-of-date image pair from
      the MVS challenge dataset. The yellow box highlights a
      region with strong shadows, varying scene contents
      such as parked cars and noise due to
      trees.}
    \label{fig:sat_pair_6}    
\end{figure}
\begin{figure*}
\centering
\includegraphics[width=0.8\textwidth]{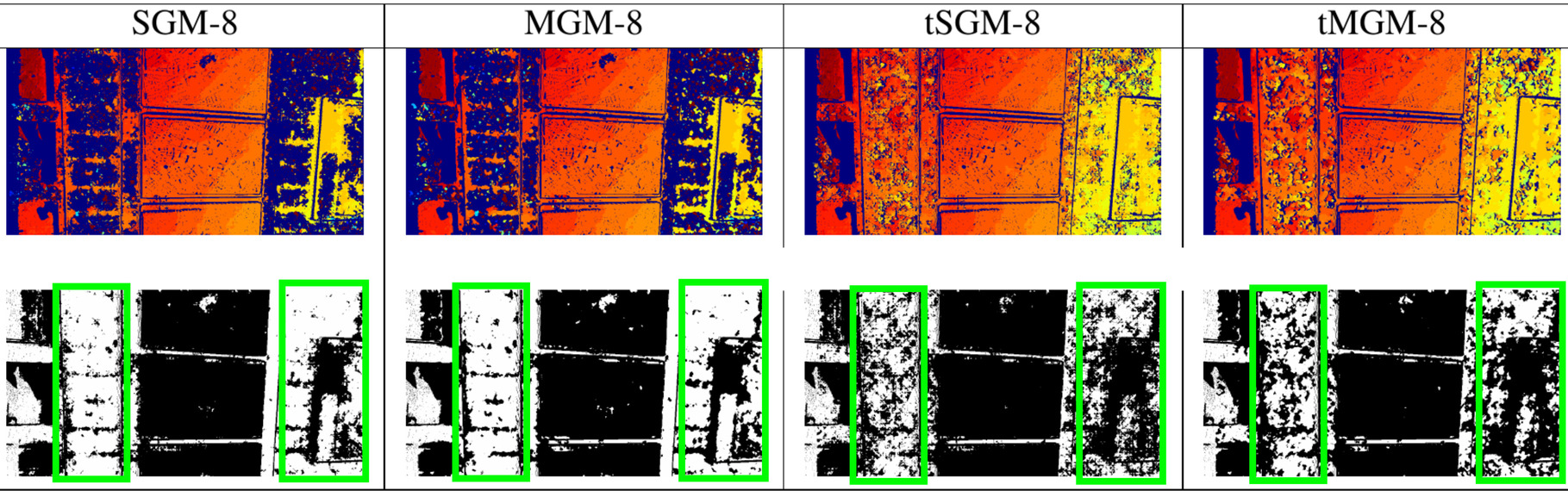}
\caption{Stereo matching results with xGM-8
  algorithms. Green boxes highlight weakly textured
  regions. The top row shows the estimated disparity maps
  and the bottom row shows the corresponding binary error
  masks, where a pixel is set to foreground if the disparity
  error $(>1)$ or if it is invalid in $\textbf{D}_b$ and
  valid in $\textbf{D}_{gt}$.}
\label{fig:untextured_xGM8}
\end{figure*}
\begin{figure*}
\centering
\includegraphics[width=0.8\textwidth]{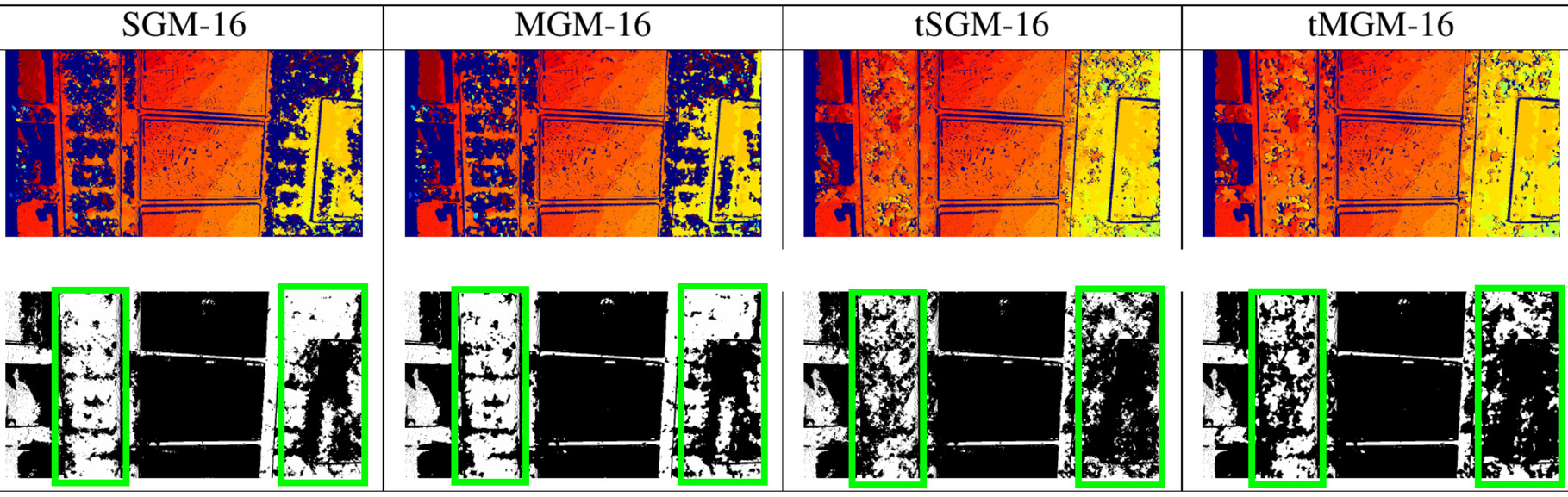}
\caption{Stereo matching results with xGM-16
  algorithms. Green boxes highlight weakly textured
  regions. The top row shows the estimated disparity maps
  and the bottom row shows the corresponding binary error
  masks, where a pixel is set to foreground if the disparity
  error $(>1)$ or if it is invalid in $\textbf{D}_b$ and
  valid in $\textbf{D}_{gt}$.}
\label{fig:untextured_xGM16}
\end{figure*}
\begin{figure*}[h]
\centering
\includegraphics[width=\textwidth]{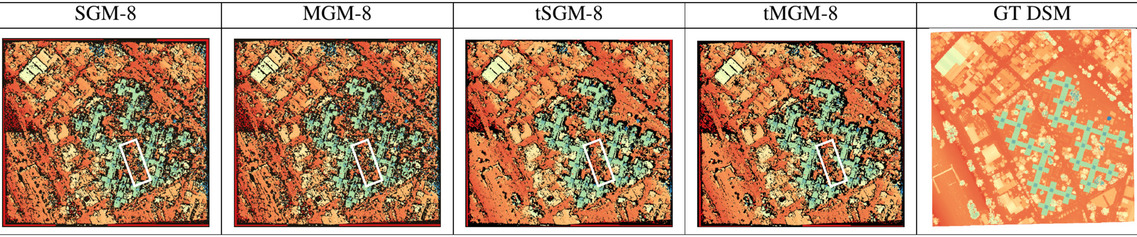}
\caption{DSM generated using xGM-8 algorithms for the
  satellite image pair from Fig. \ref{fig:sat_pair_6}. The
  white box corresponds to the yellow box in
  Fig. \ref{fig:sat_pair_6}. }
\label{fig:sat6_xGM8_results}
\end{figure*}
\begin{figure*} [h]
\centering
\includegraphics[width=\textwidth]{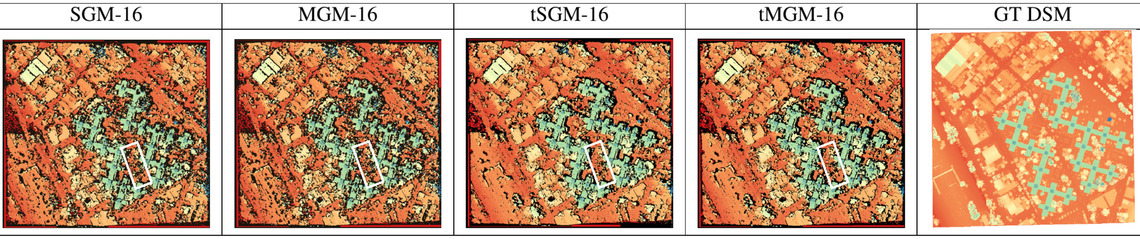}
\caption{DSM generated using xGM-16 algorithms for the
  satellite image pair from Fig. \ref{fig:sat_pair_6}. The
  white box corresponds to the yellow box in
  Fig. \ref{fig:sat_pair_6}. }
\label{fig:sat6_xGM16_results}
\end{figure*}
\subsubsection{Untextured or Weakly Textured Regions }
The ETH3D dataset mostly contains scenes with untextured or
weakly textured regions and
Fig. \ref{fig:ETH_untextured_pair} shows one such
example. The estimated disparity maps and error masks
obtained with the different algorithms are shown in
Figs. \ref{fig:untextured_xGM8} and
\ref{fig:untextured_xGM16}. In this example, the weakly
textured regions correspond to the walls and are highlighted
using green boxes. The benefits of the multi-scale census
data cost are once again observed for the tSGM and tMGM
algorithms. Moreover since the Discontinuity Cost term is
used at every level, we get denser disparity maps using the
hierarchical algorithms. The cumulative effect is that tSGM
and tMGM produce denser disparity maps compared to SGM and
MGM for untextured or weakly textured regions. For strongly
textured areas, all algorithms are comparable.



\subsubsection{Imperfect Rectification}
\label{sec:Imp_rect}
\begin{figure*}[h]
\centering
  \begin{tabular}[c]{ccc}
    \multirow{2}{*}[14pt]{
    \begin{subfigure}[ht]{0.5\textwidth}
        \includegraphics[height=4.5cm,width=\textwidth]{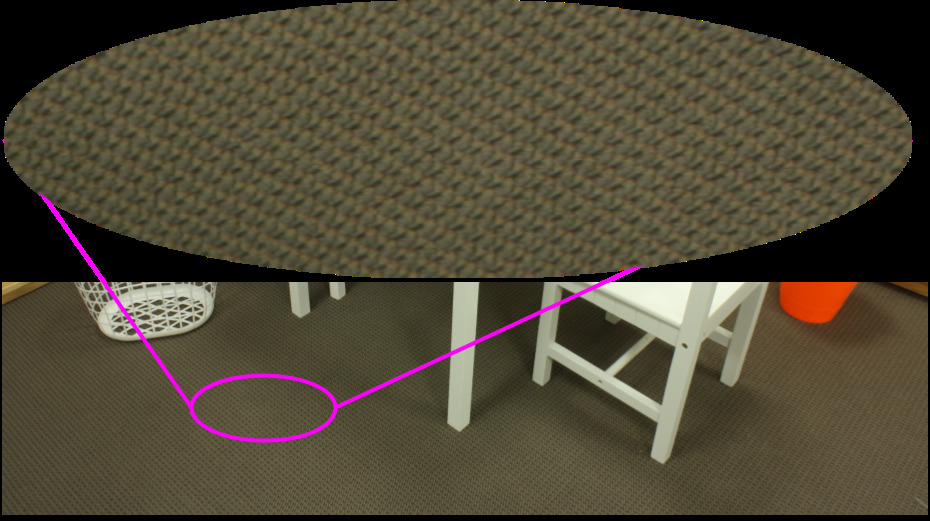}
        \caption{Left Image}
        \label{fig:repetitive_im0}
    \end{subfigure} 
    }&
    \begin{subfigure}[c]{0.18\textwidth}
        \includegraphics[width=\textwidth]{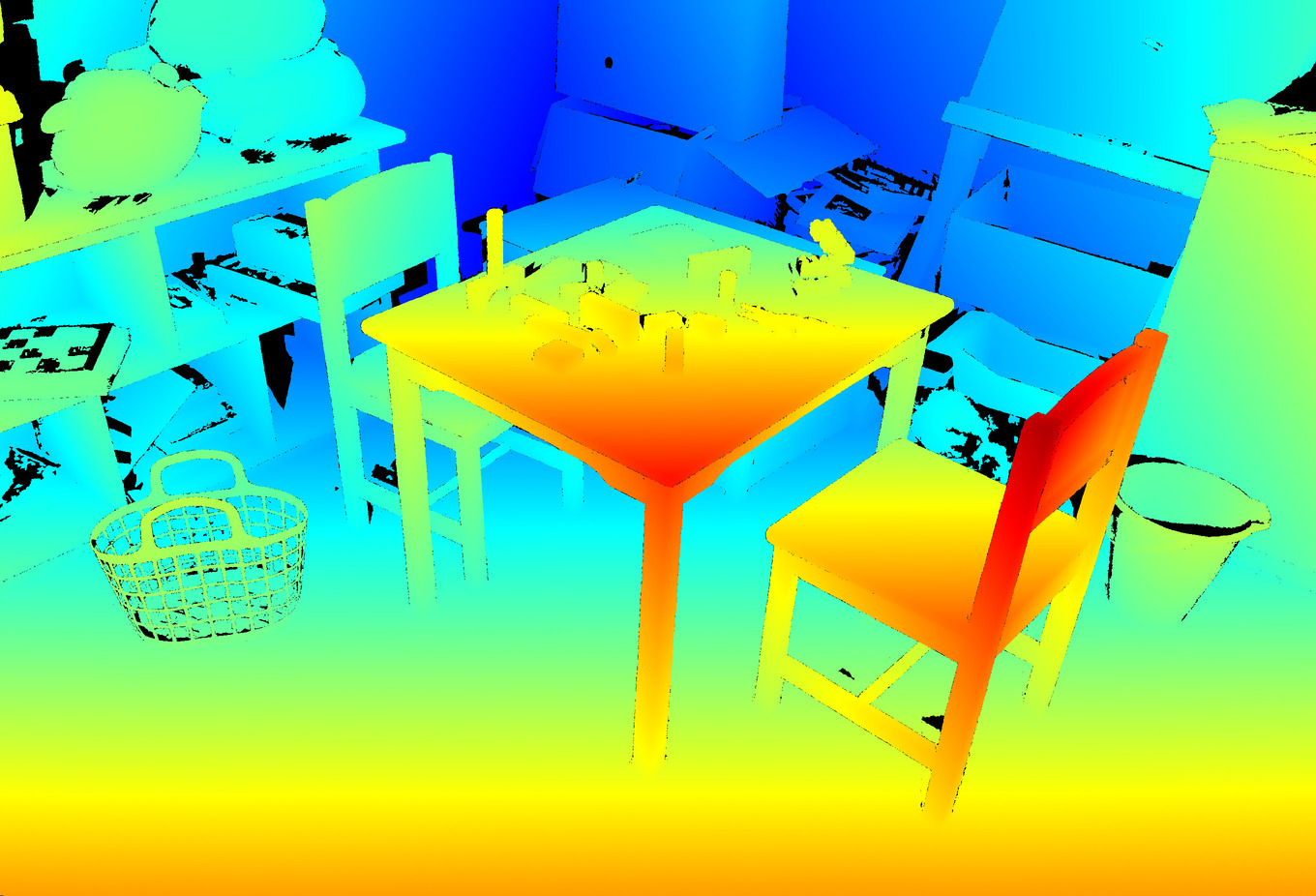}
        \caption{GT Disparity}
        \label{fig:repetitive_gt}
    \end{subfigure}\\ 
    &
        \begin{subfigure}[c]{0.18\textwidth}
        \includegraphics[width=\textwidth]{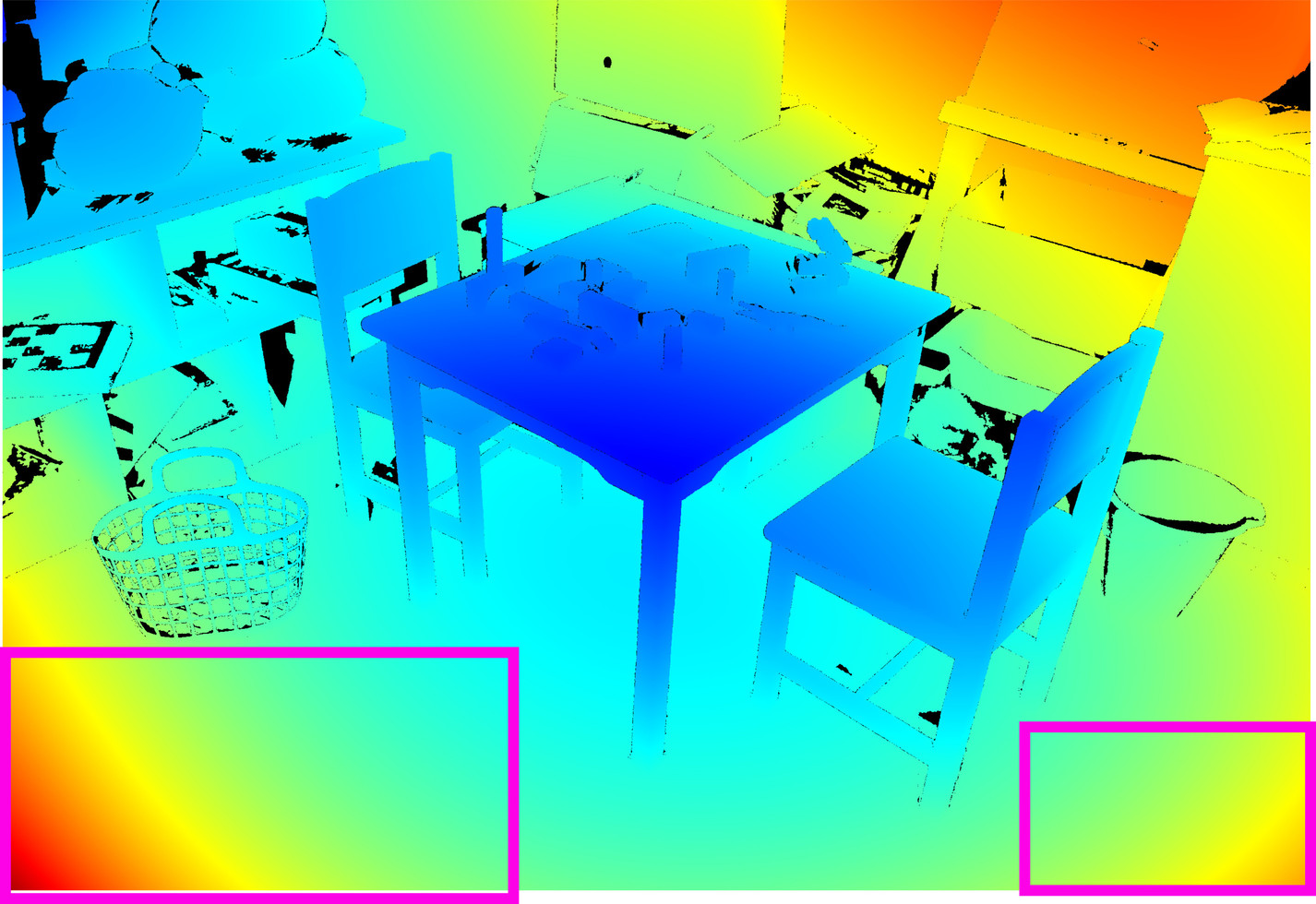}
        \caption{GT stereo rectification errors}
        \label{fig:repetitive_gt_y}
    \end{subfigure}
      \end{tabular}   \\
    \caption{An example for scenes with repetitive
      patterns. Figs. \ref{fig:repetitive_xGM8} and
      \ref{fig:repetitive_xGM16} show results using the
      xGM-8 and xGM-16 algorithms, respectively. Large
      y-alignment errors are highlighted with magenta boxes
      in \ref{fig:repetitive_gt_y}.  The y-alignment errors
      in these regions vary approximately from 0.8 to 2.6
      pixels.}
    \label{fig:repetitive_patterns}
\end{figure*}

\begin{figure*}
\centering
\includegraphics[width=0.8\textwidth]{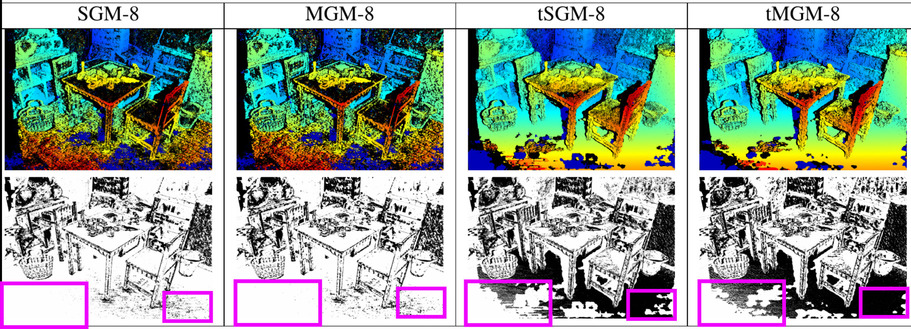}
\caption{Stereo matching results with xGM-8
  algorithms. Magenta boxes highlight areas with repetitive
  patterns. The top row shows the estimated disparity maps
  and the bottom row shows the corresponding binary error
  masks, where a pixel is set to foreground if the disparity
  error $(>1)$ or if it is invalid in $\textbf{D}_b$ and
  valid in $\textbf{D}_{gt}$.}
\label{fig:repetitive_xGM8}
\end{figure*}
\begin{figure*}
\centering
\includegraphics[width=0.8\textwidth]{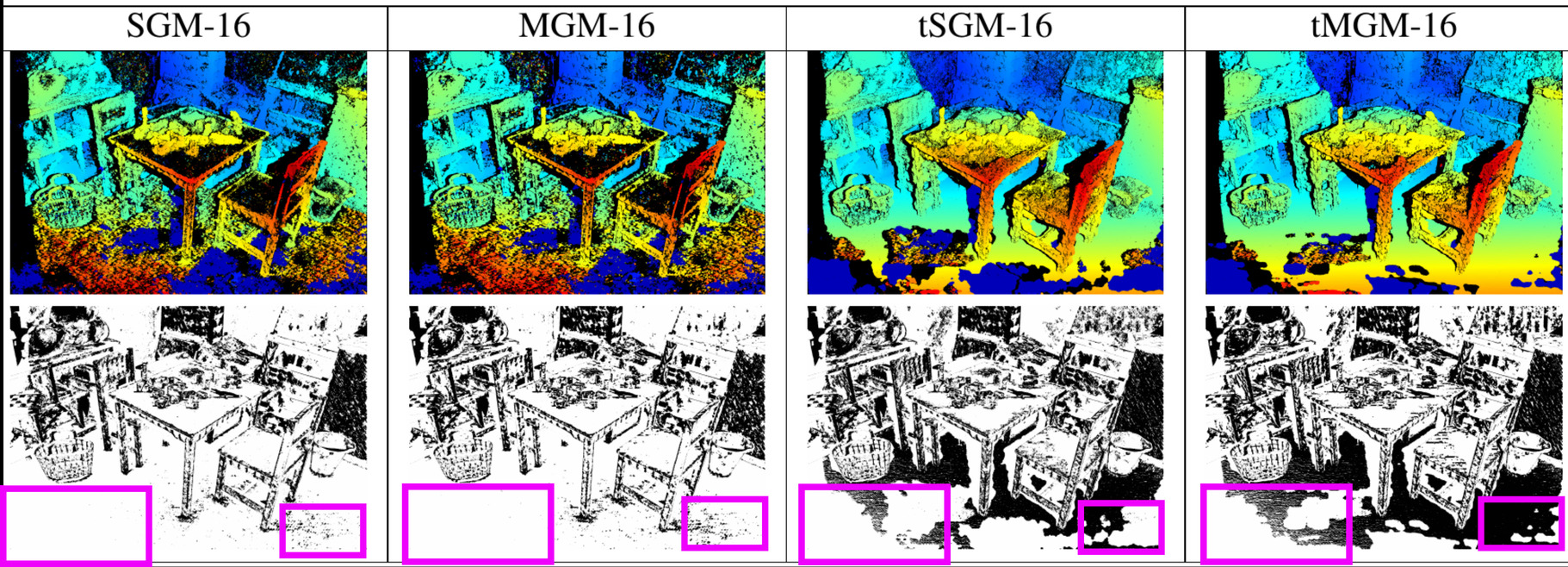}
\caption{Stereo matching results with xGM-16
  algorithms. Magenta boxes highlight areas with repetitive
  patterns. The top row shows the estimated disparity maps
  and the bottom row shows the corresponding binary error
  masks, where a pixel is set to foreground if the disparity
  error $(>1)$ or if it is invalid in $\textbf{D}_b$ and
  valid in $\textbf{D}_{gt}$.}
\label{fig:repetitive_xGM16}
\end{figure*}

In this section, we focus on the image pairs from the
Middlebury2014 dataset with imperfect rectification. Table
\ref{tbl:imp_err} reports the disparity errors averaged over
all the image pairs with imperfect rectification (with and
without illumination or exposure variations). In this table,
\emph{MB2014(I)} corresponds to image pairs with no exposure
and illumination variations, \emph{MB2014(I\_L)} corresponds
to image pairs with illumination differences and
\emph{MB2014(I\_E)} corresponds to image pairs with exposure
variations. We observe that just like in the case with
perfect rectification, tMGM-16 produces the lowest total
errors in all cases. It's interesting to note that tMGM-8
gives the lowest average error for the MB2014(I) and
MB2014(I\_E) cases, whereas tSGM-16 gives the lowest average
error for the remaining case of MB2014(I\_L).  Nevertheless,
similar to the perfect rectification case, the average error
for tSGM-8 and tMGM-16 remain close to the lowest average
errors in all cases.

The aforementioned observations for the cases of untextured
and illumination variations with perfect rectification hold
true for the cases with imperfect rectification as
well. However, special attention must be paid to scenes with
repetitive patterns. With perfect rectification, all
algorithms perform comparably in regions with repetitive
patterns. However, rectification errors in such regions can
result in ambiguities in disparity estimation. Repetitive
patterns in images can occur at both the pixel level and
structural level. An example of the latter would be a grid
of roads in an aerial image of an urban area. We are
concerned with pixel level repetitive patterns as shown in
Fig. \ref{fig:repetitive_patterns} --- the patterns are in
the carpet on the floor as should be discernible in the
magnified view of a small portion of the carpet.

The corresponding results for xGM-8 and xGM-16 are shown in
Figs. \ref{fig:repetitive_xGM8} and
\ref{fig:repetitive_xGM16} respectively. We can see that
with imperfect rectification, the hierarchical approaches
are more tolerant to stereo rectification errors in regions
with repetitive patterns.

\begin{table*}
\centering
\begin{tabular}{p{1.7cm}|p{2.6cm}|p{1.0cm}|p{0.8cm}||p{0.8cm}p{0.8cm}|p{0.8cm}p{0.8cm}|p{0.8cm}p{0.8cm}}
\toprule
& & & & \multicolumn{2}{|c|}{$>1$ pixel error} & \multicolumn{2}{c|}{$>2$ pixel error} & \multicolumn{2}{c}{$>3$ pixel error} \\
\cline{5-10} 
Algo  & Img & Avg Disp Err (pix)  & Inv Pix Err (\%) &  Bad Pix Err (\%) &  Tot Err (\%)  & Bad Pix Err (\%) &  Tot Err (\%)  & Bad Pix Err (\%) &  Tot Err (\%)   \\ \hline
\multirow{3}{1.4cm}{SGM-8} & MB2014 (I) & 11.47 & 42.54 & \textbf{16.71} & 59.25 & \textbf{10.73} & 53.27 & \textbf{9.25} & 51.78  \\ 
 & MB2014 (I\_E) & 13.09 & 45.18 & \textbf{17.26} & 62.44 & \textbf{10.44} & 55.62 & \textbf{8.98} & 54.15  \\ 
 & MB2014 (I\_L) & 23.33 & 58.59 & \textbf{17.23} & 75.82 & \textbf{11.90} & 70.49 & \textbf{10.60} & 69.19  \\ 
\hline 
\multirow{3}{1.4cm}{MGM-8} & MB2014 (I) & 14.17 & 39.74 & 18.03 & 57.77 & 12.23 & 51.97 & 10.85 & 50.59  \\ 
 & MB2014 (I\_E) & 16.51 & 42.50 & 18.69 & 61.20 & 12.01 & 54.51 & 10.63 & 53.14  \\ 
 & MB2014 (I\_L) & 28.95 & 55.52 & 19.64 & 75.16 & 14.38 & 69.90 & 13.16 & 68.68  \\ 
\hline 
\multirow{4}{1.4cm}{SGM-16} & MB2014 (I) & 10.20 & 36.51 & 19.34 & 55.85 & 11.97 & 48.48 & 10.07 & 46.59  \\ 
 & MB2014 (I\_E) & 11.71 & 39.16 & 19.97 & 59.13 & 11.74 & 50.90 & 9.88 & 49.04  \\ 
 & MB2014 (I\_L) & 20.59 & 52.04 & 20.01 & 72.05 & 13.25 & 65.29 & 11.52 & 63.57  \\ 
\hline 
\multirow{3}{1.4cm}{MGM-16} & MB2014 (I) & 10.37 & 34.01 & 19.10 & 53.11 & 12.01 & 46.01 & 10.30 & 44.30  \\ 
 & MB2014 (I\_E) & 12.25 & 36.75 & 20.03 & 56.78 & 11.92 & 48.67 & 10.22 & 46.97  \\ 
 & MB2014 (I\_L) & 22.59 & 49.35 & 20.94 & 70.29 & 14.28 & 63.63 & 12.68 & 62.03  \\ 
\hline 
\multirow{3}{1.4cm}{tSGM-8} & MB2014 (I) & 6.83 & 25.20 & 24.19 & 49.40 & 14.60 & 39.80 & 11.51 & 36.71  \\ 
 & MB2014 (I\_E) & 7.76 & 27.01 & 25.50 & 52.51 & 14.75 & 41.76 & 11.62 & 38.63  \\ 
 & MB2014 (I\_L) & 17.50 & 37.76 & 28.60 & 66.35 & 19.47 & 57.22 & 16.46 & 54.21  \\ 
\hline 
\multirow{3}{1.4cm}{tMGM-8} & MB2014 (I) & \textbf{6.67} & 23.26 & 24.43 & 47.69 & 14.87 & 38.13 & 11.84 & 35.10  \\ 
 & MB2014 (I\_E) & \textbf{7.73} & 25.02 & 25.76 & 50.78 & 14.97 & 39.99 & 11.88 & 36.90  \\ 
 & MB2014 (I\_L) & 19.13 & 36.35 & 29.28 & 65.64 & 20.27 & 56.62 & 17.36 & 53.72  \\ 
\hline 
\multirow{4}{1.4cm}{tSGM-16} & MB2014 (I) & 7.91 & 22.19 & 26.66 & 48.86 & 16.56 & 38.75 & 13.40 & 35.59  \\ 
 & MB2014 (I\_E) & 8.90 & 23.62 & 28.17 & 51.79 & 16.89 & 40.51 & 13.70 & 37.32  \\ 
 & MB2014 (I\_L) & \textbf{17.41} & 32.39 & 32.16 & 64.56 & 22.15 & 54.54 & 18.81 & 51.20  \\ 
\hline 
\multirow{3}{1.4cm}{tMGM-16} & MB2014 (I) & 7.06 & \textbf{20.50} & 25.85 & \textbf{46.35} & 15.80 & \textbf{36.30} & 12.76 & \textbf{33.26}  \\ 
 & MB2014 (I\_E) & 7.95 & \textbf{21.87} & 27.46 & \textbf{49.32} & 16.08 & \textbf{37.95} & 12.96 & \textbf{34.83}  \\ 
 & MB2014 (I\_L) & 17.80 & \textbf{31.23} & 31.87 & \textbf{63.10} & 21.89 & \textbf{53.11} & 18.64 & \textbf{49.87}  \\ 
\hline 
 
\bottomrule
\end{tabular} 


\caption{Errors averaged over the 23 image pairs with
  groundtruth and with imperfect rectification in the
  Middlebury2014 dataset. M stands for Middlebury2014. The
  suffix "\_E" is for the image pairs with exposure
  variation between views, and the suffix "\_L" is for the
  image pairs with illumination variations. Note that bad
  pixel errors only consider pixels that are valid in both
  $\textbf{D}_{gt}$ and $\textbf{D}_b$.} 

\label{tbl:imp_err}
\end{table*}

\subsection{Results on Satellite Images}
\label{subsec:Sat}
\begin{table}[h]
\centering
  \begin{tabular}{ p{1.25cm}|p{2.6cm}|p{1.2cm}p{0.8cm}p{0.8cm}}
\toprule
Algo  & Dataset & Complete- ness (\%) & Median (m) & RMSE (m)   \\ \hline
\multirow{4}{1.4cm}{SGM-8} & Explorer & 66.80 & 0.59 & 11.24  \\ 
 & MasterProvisional1 & 65.64 & 1.22 & 7.86  \\ 
 & MasterProvisional2 & 51.33 & 2.23 & 10.66  \\ 
 & MasterProvisional3 & 35.47 & 5.90 & 20.31  \\ 
\hline 
\multirow{4}{1.4cm}{MGM-8} & Explorer & 65.31 & 0.72 & 11.71  \\ 
 & MasterProvisional1 & 65.20 & 1.22 & 8.18  \\ 
 & MasterProvisional2 & 49.16 & 3.10 & 11.36  \\ 
 & MasterProvisional3 & 34.54 & 6.09 & 20.84  \\ 
\hline 
\multirow{4}{1.4cm}{SGM-16} & Explorer & 69.49 & 0.48 & 10.41  \\ 
 & MasterProvisional1 & 67.30 & 0.62 & 7.14  \\ 
 & MasterProvisional2 & 54.60 & 1.68 & 9.80  \\ 
 & MasterProvisional3 & 37.22 & 5.61 & 19.55  \\ 
\hline 
\multirow{4}{1.4cm}{MGM-16} & Explorer & 68.58 & 0.50 & 10.76  \\ 
 & MasterProvisional1 & 67.07 & 0.62 & 7.38  \\ 
 & MasterProvisional2 & 53.68 & 1.89 & 10.17  \\ 
 & MasterProvisional3 & 36.70 & 5.74 & 19.97  \\ 
\hline 
\multirow{4}{1.4cm}{tSGM-8} & Explorer & \textbf{70.93} & \textbf{0.46} & 8.66  \\ 
 & MasterProvisional1 & \textbf{67.38} & \textbf{0.51} & 6.11  \\ 
 & MasterProvisional2 & 58.43 & 1.08 & 7.22  \\ 
 & MasterProvisional3 & 38.47 & 5.11 & 18.14  \\ 
\hline 
\multirow{4}{1.4cm}{tMGM-8} & Explorer & 70.57 & \textbf{0.46} & 8.86  \\ 
 & MasterProvisional1 & 67.27 & \textbf{0.51} & 6.32  \\ 
 & MasterProvisional2 & 58.23 & 1.08 & 7.41  \\ 
 & MasterProvisional3 & 38.09 & 5.21 & 18.54  \\ 
\hline 
\multirow{4}{1.6cm}{tSGM-16} & Explorer & 70.89 & \textbf{0.46} & \textbf{8.43}  \\ 
 & MasterProvisional1 & 67.17 & \textbf{0.51} & \textbf{5.97}  \\ 
 & MasterProvisional2 & 58.60 & \textbf{1.06} & \textbf{7.05}  \\ 
 & MasterProvisional3 & \textbf{38.79} & \textbf{5.03} & \textbf{17.78}  \\ 
\hline 
\multirow{3}{1.6cm}{tMGM-16} & Explorer & 70.90 & \textbf{0.46} & 8.53  \\ 
 & MasterProvisional1 & 67.23 & \textbf{0.51} & 6.11  \\ 
 & MasterProvisional2 & \textbf{58.74} & \textbf{1.06} & 7.14  \\ 
 & MasterProvisional3 & 38.53 & 5.09 & 18.06  \\ 
\hline 
 
\end{tabular} 
\caption{Completeness, median error and RMSE averaged over 25 satellite image pairs from each region}
\label{tbl:sat_avg_err}
\end{table} 
3D stereo reconstruction from multi-date satellite images is
quite challenging because out-of-date stereo pairs can vary
significantly in illumination, scene content, sun angle,
weather conditions etc. To understand how the different SGM
variants perform under such conditions, we evaluate our
comparison on 50 satellite images collected over the San
Fernando area. These WorldView-3 images were originally
provided as part of the MVS Challenge \cite{MVS_Challenge}
and are now available from the SpaceNet dataset
\cite{spacenet}.  An example pair of images is shown in
Fig. \ref{fig:sat_pair_6}. For evaluation purposes, we focus
on four areas of interest -- Explorer, MasterProvisional1,
MasterProvisional2 and MasterProvisional3 covering areas of
0.45 sq.km, 0.13 sq.km, 0.14 sq.km and 0.1 sq.km
respectively.

Since this dataset was intended for 3D reconstruction,
ground truth disparity maps are unavailable. However, LiDAR
ground truth is provided. Therefore for a quantitative
evaluation, we generate DSMs for each SGM variant and then
use the completeness, median error and RMSE metrics defined
in \cite{MVS_Challenge}. We borrowed modules for
rectification and triangulation from the s2p pipeline
\cite{s2p} and plugged in our own implementations of the SGM
variants to construct DSMs. Table \ref{tbl:sat_avg_err}
reports errors using three metrics namely
\emph{completeness}, \emph{median error} and
\emph{RMSE}. The completeness metric measures the fraction
of 3D points with Z (height) error less than 1 meter. Higher
completeness score is desirable. For reporting median and
RMSE errors, we only consider points that are valid in
both the ground truth and generated DSMs.

From Table \ref{tbl:sat_avg_err}, we observe that tMGM-16
and tSGM-16 outperform the other algorithms on all regions
in terms of the median error and RMSE. With regard to
completeness, tMGM-16 has the best performance on
MasterProvisional2 region and produces results comparable to
the maximum on the other regions. Note that we do not fuse
the pairwise DSMs but rather compare each pairwise DSM with
LiDAR ground truth. The metrics are averaged over 25
pairs. Therefore one cannot expect our metrics to match the
results in \cite{MVS_Challenge} which are computed using
dense fused DSMs. As a future work, we would like to
investigate how these metrics relate to the Middlebury V3
metrics that directly measure error in disparities.

Figs. \ref{fig:sat6_xGM8_results} and
\ref{fig:sat6_xGM16_results} show reconstructed DSMs for the
image pair in Fig. \ref{fig:sat_pair_6}. While the DSMs
produced by SGM and MGM are noisy, tSGM-16 and tMGM-16
produce denser and cleaner DSMs even in the untextured and
shadowy regions. In order to get denser and more accurate
DSMs, one would need to fuse multiple such DSMs obtained
from different stereo pairs.

\section{Enhancements to tMGM-16 for the Middlebury Benchmark V3 Evaluation}
\label{sec:tMGM-16-Refined}
\begin{figure}[h]
 \includegraphics[width=\linewidth]{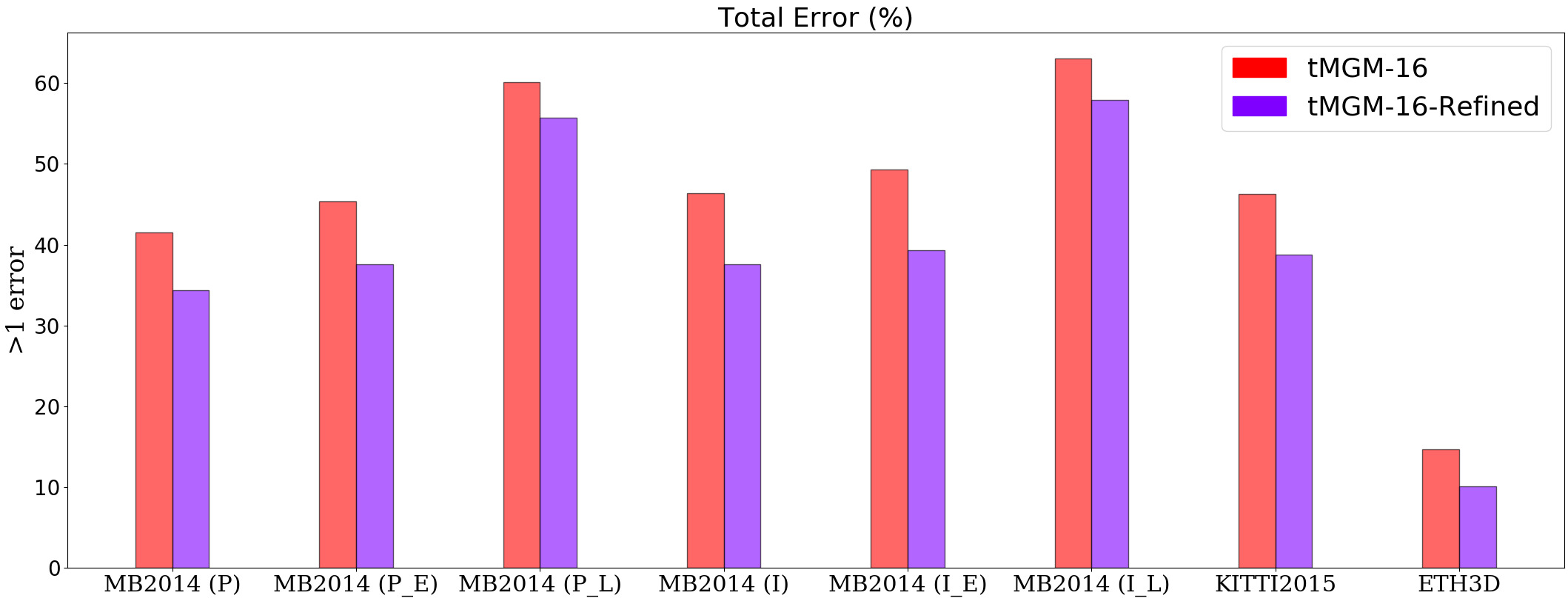} 
  \caption{Total error for $\delta > 1$ for tMGM-16 with and without the improvements described in Section \ref{sec:tMGM-16-Refined}}  
  \label{fig:tMGM-16-Refined}
\end{figure}

The discussion so far has focused on comparing the
different variants of SGM on the basis of whether or not the
implementation was hierarchical, and the type of the support
structure used.  For this comparison, we intentionally did
not apply any special post-processing to the output
disparities.  Our concern was that post-processing of any
kind could mask the consequences of the choices made for the
comparisons.

Having established the superiority of tMGM-16 for the
Middlebury2014 and the ETH3D datasets, our goal in this
section is to add enhancements to this algorithm in order to
further improve its performance.  These improvements
include:

\begin{itemize}

\item post-processing for peak removal and sub-pixel
  refinement at every level of the hierarchy

\item replacing the median filter with a joint bilateral filter 

\item a more robust way to set the disparity search range
  for valid pixels

\item new formulas for the weights $P_1$ and $P_2$

\item and, finally, dealing with the holes in the final
  disparity map.

\end{itemize}

For the sub-pixel refinement we fit a parabola to the
aggregated cost function at $\{-1,0,+1\}$ disparity values
and the location of the minima for this parabola yields a
sub-pixel shift.  The peak removal is carried out with the
same logic as presented in \cite{hirschmuller2008stereo}.

For updating the weights $P_1$ and $P_2$ we have extended
the approach presented in \cite{zbontar2016stereo} as we now
explain.  We first define the intensity differences $D_1$
and $D_2$ in the base and the match images as follows:

Let $D_1 = |I_{b}(\textbf{p})-I_{b}(\textbf{p}-\textbf{r})|$
and $D_2 =
|I_{m}(\textbf{p}+d)-I_{m}(\textbf{p}+d-\textbf{r})|$

These are then used to set the values for the weights $P_1$
and $P_2$ according to the following conditional logic that
depends on the direction of the path in the support
structure.  For pixels in the horizontal paths, $P_1$ and
$P_2$ are changed using the formula:

\[
(P_1, P_2) = \left\{
\begin{array}{l l l}
(P_1, P_2) & if \quad D_1 \leq \tau ~~ \text{and} ~~ D_2 \leq \tau \\
(P_1 / Q_2, P_2/ Q_2) & if \quad D_1 \geq \tau ~~ \text{and} ~~ D_2 \geq \tau \\
(P_1 / Q_1, P_2/ Q_1) &  \quad otherwise\\
\end{array} \right. \]

where $Q_2>Q_1$. On the other hand, for pixels in the
vertical paths, following the recommendation made in
\cite{zbontar2016stereo}, after computing $P_1$ and $P_2$
according to the formula shown above, we set $P_1 = P_1/V$.
And, when a pixel is on a diagonal path in the support
structure, we set $P_1 = P_1* \frac{\sqrt{1.0+V^2}}{V}$. In
all our experiments we have set $V = 1.25$, $\tau = 20$,
$Q_1=3$, and $Q_2 = 6$.

That brings us to the issue of how to set the search bounds
robustly for disparity estimation. Since disparity
estimation at coarser levels can be noisy, instead of taking
the minimum and the maximum as the search bounds at valid
pixels as explained in Eq. (\ref{eq:tsgm_valid}), we take
the 5th percentile as the lower bound and the 95th
percentile as the upper bound within a $41 \times 41$ window
centered at each valid pixel. 

Finally, we use discontinuity preserving interpolation
\cite{hirschmuller2008stereo} to fill holes in the final
disparity map.

In addition to the post-processing refinements described
above, we also correct for the stereo rectification errors
since, according to the Middlebury benchmark website, a
majority of the image pairs in the test dataset possess
stereo rectification errors.\footnote{It is a common
  practice to correct the vertical misalignments between the
  images that arise from camera calibration and other errors
  before submitting algorithms to the Middlebury V3
  benchmark \cite{LPS}, \cite{SPS}.}  The stereo
rectification errors are corrected by using RANSAC to fit a
linear model to the coordinates of the correspondences
between the key points extracted from the two images. The
linear model, in turn, yields the offset that brings the two
images into vertical alignment.  This step is carried out as
a pre-processing step.

Figure \ref{fig:tMGM-16-Refined} shows the effect of all the above mentioned
improvements to tMGM-16 with respect to just one metric ---
the total error for $\delta > 1$ as defined in Section
\ref{sec:Results}.

\subsection{Baseline SGM versus tMGM-16 in the Middlebury Benchmark V3 Evaluation}

The tMGM-16, refined as discussed above, was submitted for
evaluation to the Middlebury Benchmark V3.  The benchmark
returned the results shown in Table \ref{tbl:middle_V3}.
The column headings are the $\delta$ values for which the
weighted average errors were computed using 15 test image
pairs. The benchmark computes a large number of evaluation
results using different metrics. Table \ref{tbl:middle_V3}
shows the results for the total errors at the non-occluded
pixels.  The first row of the results are for the baseline
SGM as submitted by Hirschmuller
\cite{hirschmuller2008stereo}.  As can be seen in the table,
tMGM-16 achieves an improvement of 6-8\% on the average over
the baseline SGM.

\begin{table}[h]
\centering
\begin{tabular}{p{2.5cm} | p{1.0cm}p{1.0cm}p{1.0cm}p{1.0cm}}
\hline
Method & 0.5px & 1px & 2px & 4px \\
\hline
SGM-HH & 55.0     & 35.8   & 25.3   & 19.0 \\
tMGM-16 & 48.2 & 27.8 & 17.3 & 11.5 \\
\hline
\end{tabular}
\caption{Official results for non-occluded pixels from the
  Middlebury benchmark. SGM-HH results are for the optimized
  baseline SGM algorithm as submitted by
  \cite{hirschmuller2008stereo}. }
\label{tbl:middle_V3}
\end{table}

\section{DEM-Sculpting for Satellite Stereo Matching}

\begin{figure*}[ht]
\centering
    \begin{subfigure}[b]{0.45\textwidth}
        \includegraphics[width=\textwidth, height=70mm]{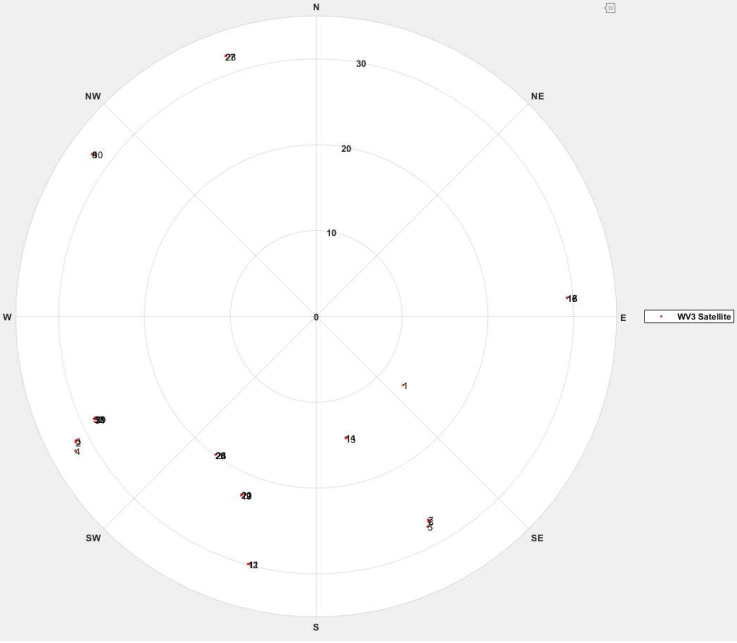}
        \caption{Azimuth and obliquity angle distribution of satellite camera.}
        \label{fig:sat_angles}
    \end{subfigure}
    ~ 
    \begin{subfigure}[b]{0.45\textwidth}
        \includegraphics[width=\textwidth, height=70mm]{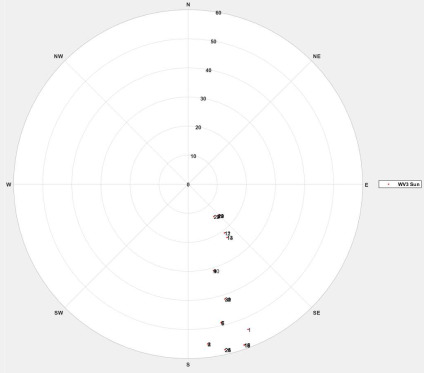}
        \caption{Azimuth and obliquity angle distribution of sun.}
        \label{fig:sun_angles}
    \end{subfigure}
    \caption{Distribution of azimuth and obliquity angles of satellite
and sun. Red dot represents each WV3 image out of thirteen unique images
captured in Kabul region.  }
    \label{fig:kabul_dist}
  \end{figure*}

\begin{figure*}[ht]
\centering
        \includegraphics[width=\textwidth, height=60mm]{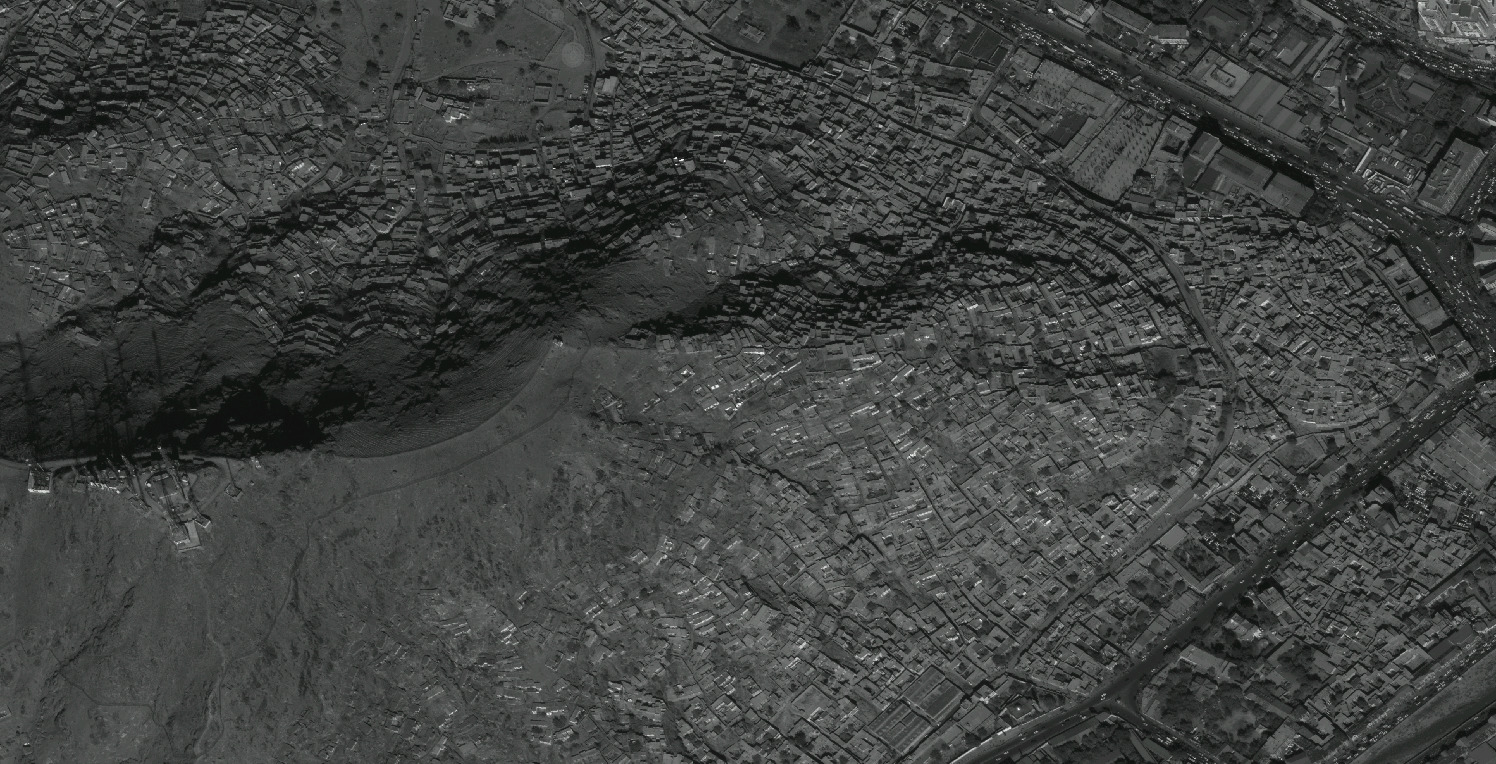}
    \caption{A WV3 image showing the mountain region in Kabul.}
    \label{fig:kabul_satimg}
  \end{figure*}

  In this Section, we present a case study of stereo reconstruction
using a few multi-date satellite images that are acquired under the
limited range of azimuth and obliquity angles of a satellite as well
as the sun. We fixed the stereo matching algorithm as tSGM-8. Fig.
 \ref{fig:kabul_dist} shows the plots for satellite and sun angle distribution of the
available number of WV3 images. The outer circle marked with
‘E’, ‘NE’ and so on represents azimuth angle and the concentric
circles represent the obliquity angle. Note that from the plot of
sun angle distribution, all the images are captured around the sun
rise time and the obliquity angles are very large that means the
elevation angles are very low. Additionally, the region contains
a tall mountain with small mud houses on its elevation which introduces
additional stereo matching challenges due to large disparity search
range and untextured regions (see Fig. \ref{fig:kabul_satimg}). Out of thirteen unique
WV3 images two were heavily occluded due to clouds. It’s widely
known that a wider baseline yields poor results in stereo matching
and a very narrow baseline yields poor results in triangulation.
Therefore, additional thresholds are applied to filter out stereo
pairs based on view-angle and acquisition time differences. After
all the pre-processing, only 17 stereo pairs were available to
generate the fused DSM for the Area of Interest (AOI) in Kabul.

\begin{figure*}[ht]
\centering
    \begin{subfigure}[b]{0.30\textwidth}
        \includegraphics[width=\textwidth, height=40mm]{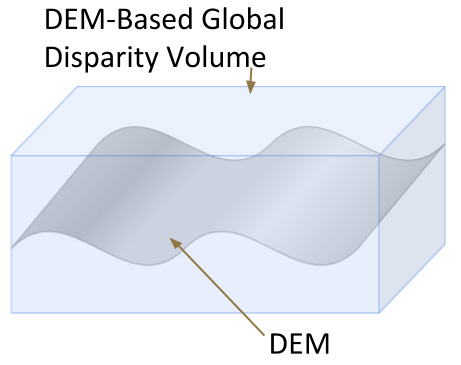}
        \caption{DEM-based global disparity search bound.}
        \label{fig:dem_global}
    \end{subfigure}
    ~ 
    \begin{subfigure}[b]{0.60\textwidth}
        \includegraphics[width=\textwidth, height=40mm]{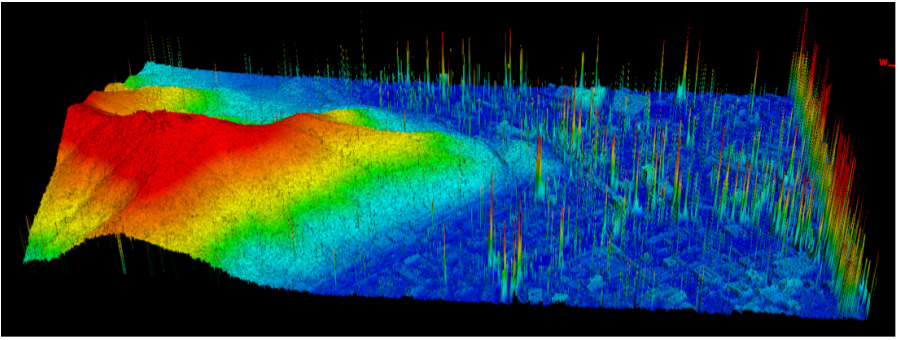}
        \caption{Kabul fused DSM (DEM-based global disparity search bounds)}
        \label{fig:dem_global_DSM}
      \end{subfigure}
      ~
          \begin{subfigure}[b]{0.30\textwidth}
        \includegraphics[width=\textwidth, height=40mm]{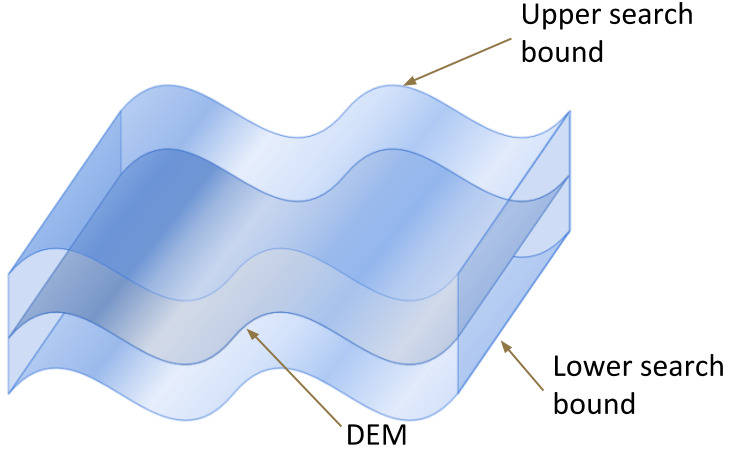}
        \caption{DEM-sculpted disparity search bound.}
        \label{fig:dem_sculpting}
    \end{subfigure}
    ~ 
    \begin{subfigure}[b]{0.60\textwidth}
        \includegraphics[width=\textwidth, height=40mm]{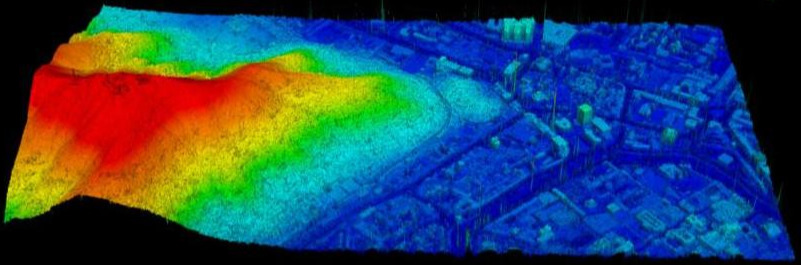}
        \caption{Kabul fused DSM (DEM-sculpted disparity search bound)}
        \label{fig:dem_sculpt_DSM}
      \end{subfigure}
    \caption{DEM-based global disparity search bounds vs DEM-sculpted disparity search bounds}
    \label{fig:kabul_dem_before_after}
  \end{figure*}

A common practice to estimate the initial global disparity search
bound is to use a set of world points sampled either on a 3D grid
generated using the extents available in Rational Polynomial
Coefficient (RPC) \cite{Dial2005} camera model \footnote{In general,
  satellite or pushbroom camera parameters are represented as RPC
  model.} or using a low resolution SRTM (Shuttle Radar Tomography
Mission) DEM (Digital Elevation Model), e.g., \cite{s2p}). For the rest of the
discussion, we assume the RPC parameters are corrected using a
commonly used approach, based on bundle adjustment
\cite{Triggs2000}. In other words, image-to-image alignment is
performed with sub-pixel accuracy.  Fig. \ref{fig:dem_global} shows
the intuition behind estimating global disparity search bounds using
DEM and the fused DSM output using 17 stereo pairs. Clearly, the
global search bound is too large in flat region as compared to the
mountain region causing very large spikes in the flat region.

We propose a novel DEM-sculpting logic to estimate tighter per-pixel
search bounds to alleviate the aforementioned problem. We now
summarize the algorithm in the following steps.
\begin{enumerate}
\item Add upper and lower
offsets to DEM, estimated per RPC camera model. An upper offset $u_o$
is estimated as $$u_o = 0.5(h^{RPC}_{max} - h_{max}^{DEM})$$ where
$ h_{max}^{DEM}$ is the maximum DEM height and $h^{RPC}_{max}$ is the
maximum height for a given RPC camera model. Similarly, a lower offset
is estimated as

$$l_o = 0.15(h_{min}^{RPC}- h_{min}^{DEM})$$
where $h_{min}^{RPC}$ and $h_{min}^{DEM}$ are minimum heights in RPC
and DEM, respectively. We obtain the two DEMs after adding $u_o$ and $l_o$
to the original DEM (see Fig. \ref{fig:dem_sculpt_DSM}).

\item Generate a set of sparse points from a 2D grid in stereo
rectified reference image.

  
\item Map the sampled points into the corresponding reference
unrectified view. Note that we use the approach proposed by Oh \etal
\cite{oh2010piecewise} to stereo rectify the satellite images. The
approach by Oh \etal produces a rectification $xy$-map which maps
point locations in rectified view to the corresponding unrectified
view. We approximate the inverse mapping using a KDTree-based nearest
neighbor search and inverse distance weighted interpolation.

\item Backproject the sampled points from the unrectified reference
view to the DEMs obtained by adding $u_o$ and $l_o$ to obtain two sets
of world points.

\item Forward project the world points to the unrectified secondary
view and then mapped to the rectified secondary view using the inverse
mapping as explained in Step (3).

\item Steps 1-5 give us the sparse correspondences in a reference and
the secondary rectified views. By applying the definition of disparity
we get lower and upper disparity search bounds.

 \item The holes in sparse search bound maps are then filled using
the nearest neighbor interpolation.
\end{enumerate}

Note that in theory, one could
  start with all the points in a reference rectified views, however
  the backprojection using RPC is computationally expensive. We
  noticed that for stereo matching purpose coarse search bound
  estimation offers a good compromise between speed and accuracy.
Fig. \ref{fig:dem_sculpt_DSM} shows the fused DSM result using
DEM-sculpting logic for the initial disparity search bounds.
\section{Conclusions}

Our results show that hierarchical approaches are not just
faster but are also more accurate than their
non-hierarchical counterparts for all datasets. This
improvement stems from using the multi-scale census transform and the
Discontinuity Cost term at every level in the hierarchy. The
performance of tMGM-16 is the best or close to the best for all
datasets except the KITTI2015 dataset. Possible reasons and solutions
to improve the performance for the KITTI2015 dataset have been
discussed in Section \ref{sec:Results}. In general, even in the
presence of significant illumination differences, untextured or weakly
textured regions, repetitive structures and imperfect rectification,
the tMGM-16 algorithm is the best to use for disparity
calculations. The combination of the hierarchical approach,
``quad-area'' based support regions and 16 directions in tMGM-16
produces dense and accurate disparity maps. We present a novel
DEM-sculpting approach to estimate initial disparity search bounds
which further supports our observations about the effects of tighter
search bounds on the overall matching accuracy.


\ifCLASSOPTIONcompsoc
  \section*{Acknowledgments}
\else
  \section*{Acknowledgment}
\fi
Supported by the Intelligence Advanced Research Projects
Activity (IARPA) via Department of Interior / Interior
Business Center (DOI/IBC) contract number D17PC00280. The
U.S. Government is authorized to reproduce and distribute
reprints for Governmental purposes notwithstanding any
copyright annotation thereon. Disclaimer: The views and
conclusions contained herein are those of the authors and
should not be interpreted as necessarily representing the
official policies or endorsements, either expressed or
implied, of IARPA, DOI/IBC, or the U.S. Government.

{\small \bibliographystyle{ieee}
  \bibliography{egbib} }

%
%
%
%
\end{document}